\PassOptionsToPackage{clashing-option}{package-name}
\PassOptionsToPackage{mainaux}{rerunfilecheck}
\documentclass[dvipsnames,format=sigconf,authorversion, nonacm, preprint]{acmart}
\renewcommand\footnotetextcopyrightpermission[1]{} 
\setcopyright{none} 

\usepackage{algorithm}
\usepackage{algpseudocode}
\usepackage{multirow}
\usepackage{booktabs}
\usepackage{colortbl}
\usepackage[table]{xcolor} 
\usepackage{pgf}           
\usepackage{makecell}      
\definecolor{tie}{gray}{0.9} 

\definecolor{tie}{RGB}{255,235,190}

\newcommand{\gradcell}[4]{%
  \ifnum#2=#3%
    \cellcolor{tie}\makecell{#1/#2/#3}%
  \else%
    \pgfmathtruncatemacro{\diff}{abs(#2-#3)}%
    \pgfmathtruncatemacro{\shade}{min(100, round(100*\diff/#4))}%
    \ifnum#2>#3%
      \edef\temp{\noexpand\cellcolor{red!\shade}}\temp%
    \else%
      \edef\temp{\noexpand\cellcolor{green!\shade}}\temp%
    \fi%
    \makecell{#1/#2/#3}%
  \fi%
}

\begin{document}

\title{Search-Based Risk Feature Discovery in Document Structure Spaces under a Constrained Budget}


\author{Saisubramaniam Gopalakrishnan, Harikrishnan P M, Dagnachew Birru}
\affiliation{%
  \institution{Philabs, Quantiphi}
  \country{India, USA}
}
\email{{gopalakrishnan.saisubramaniam, harikrishnan.m, dagnachew.birru}@quantiphi.com}


\begin{abstract}
Enterprise-grade Intelligent Document Processing (IDP) systems support high-stakes workflows across finance, insurance, and healthcare. Early-phase system validation under limited budgets mandates uncovering diverse failure mechanisms, rather than identifying a single worst-case document. 
We formalize this challenge as a Search-Based Software Testing (SBST) problem, aiming to identify complex interactions between document variables, with the objective to maximize the number of distinct failure types discovered within a fixed evaluation budget. Our methodology operates on a combinatorial space of document configurations, rendering instances of structural \emph{risk features} to induce realistic failure conditions. We benchmark a diverse portfolio of search strategies spanning evolutionary, swarm-based, quality-diversity, learning-based, and quantum under identical budget constraints.
Through configuration-level exclusivity, win-rate, and cross-temporal overlap analyses, we show that different solvers consistently uncover failure modes that remain undiscovered by specific alternatives at comparable budgets. Crucially, cross-temporal analysis reveals persistent solver-specific discoveries across all evaluated budgets, with no single strategy exhibiting absolute dominance. While the union of all solvers eventually recovers the observed failure space, reliance on any individual method systematically delays the discovery of important risks. These results demonstrate intrinsic solver complementarity and motivate portfolio-based SBST strategies for robust industrial IDP validation.

\end{abstract}

\fancypagestyle{firstpage}{
  \fancyhf{}
  \renewcommand{\headrulewidth}{1pt}      
  \renewcommand{\footrulewidth}{0pt}    
  \fancyfoot[L]{\\\textit{Preprint. Under Review.}}
}
\maketitle
\thispagestyle{firstpage} 
\section{Introduction}

The operational landscape of modern enterprises increasingly relies on Intelligent Document Processing (IDP) systems to automate high-volume decision-making workflows across finance, insurance, healthcare, and regulatory domains. Document AI \cite{cui2021document} has emerged as a central research and industrial pursuit for automatically reading, understanding, and extracting information from complex, visually-rich business documents, including receipts, invoices, insurance policies, tax forms, and financial summaries. Despite substantial progress on academic benchmarks for layout analysis, entity extraction, and document classification, real-world deployments often expose significant reliability gaps: models that achieve strong accuracy on curated test sets nonetheless exhibit brittle performance when encountering structural variability and domain drift that occur in practice \cite{cui2021document, google2023advances}.

Standard aggregate performance metrics such as accuracy or F1-score hide failure modes triggered by specific combinations of layout patterns, noise, and pagination that are under-represented in static test sets. Traditional robustness testing using adversarial perturbations at the pixel level \cite{pintore2024counterfeit}, while theoretically valuable, target imperceptible changes that are rarely reflective of the semantic and structural risk features that dominate failures in enterprise document pipelines. Moreover, systems based on deep learning models that are deployed in production frequently encounter distribution drifts (development-to-operation shift) \cite{zohdinasab2023deepatash}, resulting in degraded performance and silent failures in downstream tasks such as information extraction and compliance checking.

In this work, we address the IDP validation challenge through the lens of \emph{Search-Based Software Testing (SBST)} \cite{harman2001sbse, afzal2009systematic, mcminn2011search}, framing the problem as \emph{budgeted risk feature discovery}. Rather than searching for a single document that induces worst-case behavior, the objective is to systematically uncover a diverse set of distinct failure mechanisms under a constrained evaluation budget. We operationalize this as exploration over a structured risk feature space describing document configurations, which are rendered into synthetic documents using a generator and evaluated at the system level by an IDP oracle to produce structured failure signatures. This formulation shifts validation from scalar optimization toward coverage and diversity of failure behavior, aligning with diversity-oriented and feature-space testing approaches for learning-based systems \cite{zohdinasab2021deephyperion}, and addressing a critical gap in enterprise-grade Document AI validation. Our contributions are as follows:
\paragraph{\textbf{Risk Feature Discovery as Search-Based Software Testing:}}
We formulate validation of IDP systems as a Search-Based Software Testing (SBST) problem over structured document configuration spaces. Our formulation targets the budget-constrained discovery of \emph{diverse risk features / failure mechanisms}, directly reflecting the realities of enterprise-scale IDP validation.
\paragraph{\textbf{A Budget-Matched, System-Level Benchmark for Risk Discovery:} }
We introduce a controlled benchmark for evaluating heterogeneous search strategies under identical evaluation budgets and a shared system-level IDP oracle. Performance is assessed using complementary metrics that capture both risk severity and discovery behavior, including top-decile risk statistics and method-wise discovery overlap. This enables fair comparison across evolutionary, learning-based, quality–diversity, and quantum-inspired solvers beyond single-metric optimization.
\paragraph{\textbf{Structural and Cross-Temporal Evidence of Solver Complementarity:} } 
We propose a pairwise exclusivity and cross-temporal overlap analysis that quantifies each solver’s marginal contribution relative to others at matched and mismatched budgets. Across document configuration spaces, different methods exhibit persistent temporal advantages and structural biases, discovering distinct failure signatures earlier or maintaining exclusive discoveries over extended budget horizons. These results demonstrate solver complementarity under realistic testing constraints, even in the absence of a universally dominant method.
\paragraph{\textbf{Risk Mode Characterization and Predictive Modeling:}  }
We introduce a discrete abstraction of observed failures into \emph{risk modes}, capturing recurring structural patterns and failure mechanisms induced by document configurations. Using the accumulated evaluation data, we further demonstrate that future document risk can be accurately predicted (\(R^2 > 0.91\)) using lightweight models such as Random Forest. This enables anticipatory risk assessment and supports proactive test prioritization in enterprise IDP validation.

\section{Related Work}

\subsection{Software Testing for Document Processing}
IDP evaluation transcends traditional software testing as failures arise from the non-linear coupling of spatial geometry and semantic structure. Robustness depends on how the interpretation of visual hierarchies in layouts mediates the transition from raw pixels to structured table schemas and key-value associations, where localized geometric distortions can propagate into systemic extraction errors.
Modern IDP pipelines are predominantly built on Deep Learning (DL) models and operate as black-box systems with expensive evaluation costs, making exhaustive testing infeasible. This has motivated the use of \emph{Search-Based Software Testing (SBST)} \cite{harman2001sbse, mcminn2011search}, which formulates test generation as an optimization problem under constrained budgets. Beyond Combinatorial Interaction Testing \cite{petke2015practical}, recent work has begun to adapt SBST principles to DL, emphasizing feature-space exploration to uncover diverse errors beyond what is observed on static test sets \cite{zohdinasab2021deephyperion}. These ideas are less explored in systematic IDP validation under constrained budgets, motivating search-based exploration of realistic risks.

\subsubsection{OCR and Textual Key-Value Extraction:}
Early IDP pipelines relied on modular OCR engines \cite{smith2007tesseract}, with robustness primarily evaluated using character error rate (CER) under image degradations. Modern architectures such as Donut \cite{kim2021donut} and Pix2Struct \cite{lee2023pix2struct} integrate visual and textual processing end-to-end, reducing reliance on standalone OCR. While improving contextual reasoning, this integration introduces failure modes where visual artifacts or layout noise are misinterpreted as valid text \cite{pintore2024counterfeit}. Such semantic textual errors can silently affect downstream tasks; for example, corrupted text regions in clinical documents can alter classification outcomes without obvious visual cues \cite{fatehi2023towards}. In contrast to optimizing CER or generating imperceptible adversarial perturbations, our focus is on realistic OCR-induced risks that propagate through document structure and layout into system-level failures.

Key–Value (KV) extraction is another core capability of enterprise IDP systems, requiring joint reasoning over textual content and spatial layout, as exemplified by LayoutLM-style models \cite{huang2022layoutlmv3}. While benchmarks such as FUNSD \cite{jaume2019funsd} and SROIE \cite{huang2019icdar2019} report field-level accuracy, KV extraction is sensitive to layout density, spatial misalignment, and field interactions. These factors are typically implicit in datasets rather than systematically explored, motivating budgeted search over explicit KV risk features.

\subsubsection{Document Layout and Structural Verification:}
Beyond text, IDP systems must accurately parse document layout into logical regions such as headers, paragraphs, tables, and footers. Prior work evaluates layout models using benchmark datasets and perturbation-based robustness studies, showing sensitivity to spacing, alignment, and region density \cite{zhang2024robustness}. Adversarial attacks based on predefined perturbation suites further demonstrate that small geometric shifts can cause cascading segmentation failures \cite{chen2024rodla}. Recent approaches generate document layouts using structured representations such as scene graphs for layout, enabling controllable synthesis of complex layouts and interactions between regions \cite{8948239, 10.1007/978-3-030-86334-0_36}, while LLM-based methods extend this paradigm toward holistic document generation by jointly synthesizing spatial structure and region-level content \cite{ke2025large, harikrishnan2025docgenie} and typically used for augmentation than part of validation.

\subsubsection{Tabular Structure and Relational Consistency:}
Table understanding is a critical and failure-prone component of IDP systems, requiring models to correctly detect tables and infer row–column relationships. Prior work has shown that table detection and structure recognition models are sensitive to layout variability and diverse table styles \cite{schreiber2017deepdesrt, gilani2017table}. Such errors can lead to silent failures where extracted values are misaligned or incorrectly associated. Existing evaluations of these methods have largely focused on performance on fixed benchmarks and comparative surveys of modeling approaches \cite{zanibbi2004survey, alexiou2023evaluation}, rather than systematic robustness under interacting structural variations. In contrast, industrial validation requires exploring interacting table configurations such as varying row/column counts, multi-page continuation, and co-occurrence with other content blocks—under constrained budgets, motivating structured search over table-centric risk features.

\subsection{Optimization Primitives for Exploration}
Effective validation under constrained budgets requires search strategies that balance sample efficiency, exploration, and coverage of diverse failure mechanisms. We briefly summarize the optimization families evaluated in this work.

\subsubsection{Evolutionary, Swarm, and Quality–Diversity Methods:}
Population based evolutionary algorithms, such as Genetic Algorithms (GA), explore rugged search spaces through stochastic variation and selection, enabling parallel exploration across multiple regions \cite{holland1975adaptation}. Swarm-based methods like Particle Swarm Optimization (PSO) guide search via shared local and global information, often yielding rapid convergence but correlated trajectories \cite{kennedy1995pso}. Multi-Objective Evolutionary Algorithms (MOEA), such as NSGA-II \cite{deb2002fast}, extend this paradigm by explicitly optimizing trade-offs among competing objectives. While powerful for balancing multiple scalar criteria, MOEAs typically require predefined objective formulations and do not directly optimize for behavioral diversity or coverage unless diversity is encoded as an explicit objective. 
Quality–Diversity (QD) \cite{pugh2016qd} methods such as MAP-Elites \cite{mouret2015mapelites} decouple performance from diversity by maintaining an archive over discretized feature dimensions, promoting coverage across structurally distinct regions of the search space. This makes them well-suited for risk feature discovery, where uncovering heterogeneous failure mechanisms is prioritized over optimizing a single (or multi-objective) performance metric.
\subsubsection{Bayesian Optimization:} BO \cite{snoek2012practical} is designed for sample-efficient optimization of expensive black-box functions using probabilistic surrogate models. Gaussian Process (GP)-based BO combines uncertainty modeling with acquisition functions such as Expected Improvement (EI) \cite{jones1998ego} and Upper Confidence Bound (UCB) \cite{srinivas2010gpucb}, and is effective in continuous, low-dimensional spaces. Tree-structured Parzen Estimators (TPE) \cite{bergstra2011tpe} extend BO to discrete and hierarchical spaces and are better suited to find optima for combinatorial domains. 
\subsubsection{Reinforcement Learning:} RL frames search as sequential decision making, learning a proposal policy from observed rewards. Policy-gradient methods such as REINFORCE \cite{williams1992reinforce} and PPO \cite{schulman2017ppo} can adapt sampling behavior over time. However, because training optimizes a scalar reward, diversity in discovered configurations typically arises indirectly from policy stochasticity or reward shaping rather than being an explicit objective.
\subsubsection{Stochastic Local Search:} Simulated Annealing (SA) \cite{kirkpatrick1983annealing} provides a simple baseline for combinatorial optimization by probabilistically accepting worse solutions to escape local optima. While the temperature schedule allows to traverse energy barriers that trap greedy search strategies, making it suitable for rugged landscapes and local exploration, SA operates sequentially and may under-explore distant regions of the search space under strict budgets.
\subsubsection{Quantum Optimization via Learned Surrogates:}
Quantum optimization methods, including Quantum Annealing \cite{kadowaki1998quantumannealing} and Variational methods such as Quantum Approximate Optimization Algorithm (QAOA) \cite{farhi2014qaoa}, have been proposed for combinatorial problems with complex interaction structures. Black-box objectives are converted into quadratic surrogates in QUBO form~\cite{glover2018qubo}, e.g., Factorization Machines (FM)~\cite{rendle2010fm} is used with quantum annealing for applications like materials design~\cite{kitai2020metamaterials}.
For smaller document structure configurations, quantum serves primarily to induce unique exploration dynamics via learned interaction and structured mixing, complementing classical heuristics under constrained budgets. By leveraging superposition, entanglement, and interference \cite{deutsch1985quantum}, it can bias sampling toward regions difficult to reach via local mutations or gradient-based search and offers a potential pathway to scalable combinatorial search as quantum hardware advances.

\section{Methodology}
\label{section:methodology}

\begin{figure}[h]
  \centering
  \includegraphics[width=0.95, width=\linewidth]{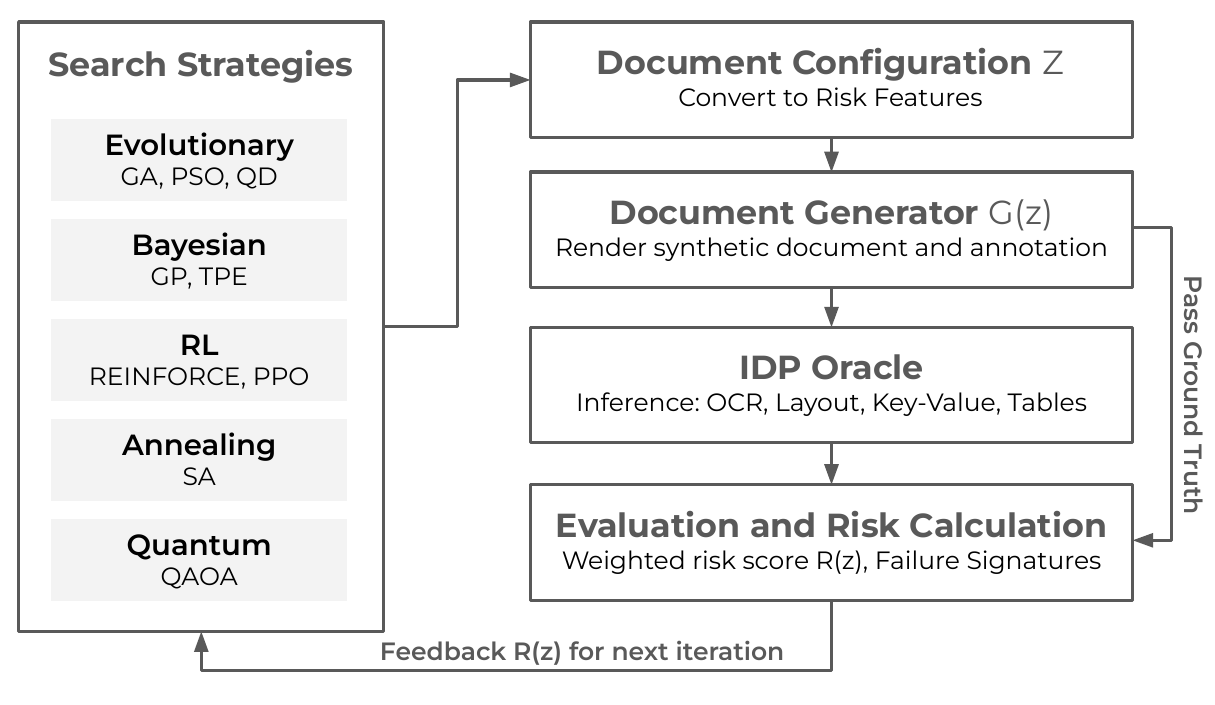}
  \caption{Schematic of the Risk Feature Discovery Pipeline}
  \label{fig:pipeline}
\end{figure}

\subsection{Problem Formulation and Risk Features}
We formulate the validation of IDP systems as a combinatorial search problem aimed at uncovering system-level weaknesses. 
Unlike traditional testing, which relies on pre-curated datasets and assumes i.i.d. conditions, we seek to identify complex interactions between document characteristics that trigger failures, thereby exposing latent blind spots that aggregate metrics on static benchmarks fail to capture. 
The Risk Feature discovery pipeline is provided in Figure \ref{fig:pipeline}. We define the document configuration space $\mathcal{Z}$ as a finite set of discrete decision variables, which we term \textit{Risk Features}. A configuration $z \in \mathcal{Z}$ serves as a structural blueprint: \[ z = (x_{\text{template}}, x_{\text{density}}, x_{\text{noise}}, x_{\text{artifacts}}, \ldots)\]
These features include categorical choices (e.g., template structure), ordinal counts (e.g., number of pages, table rows, key–value pairs, noise levels), and additional binary flags based on industry domains. Each configuration corresponds to a concrete \emph{test case} for the IDP system, specifying a complete document structure and rendering parameters.
To enable standard combinatorial optimization, we encode these variables into a unified binary vector of dimension $N$. Categorical variables are one-hot encoded, while ordinal counts are bin-mapped to binary strings. This transformation maps the problem onto the binary hypercube $\{0,1\}^N$, allowing standard bit-flip and QUBO-based search heuristics.
The cardinality of $\mathcal{Z}$ is prohibitively large for exhaustive enumeration, necessitating efficient search strategies to locate sparse, high-risk regions.

\subsection{Synthetic Document Generation}
A procedural document generator $\mathcal{G}: \mathcal{Z} \rightarrow \mathcal{I}$ maps the binary configuration $z$ to a rendered document instance $d$. Because the document is synthetically generated, $\mathcal{G}$ simultaneously produces ground-truth annotations for all downstream tasks (text sections, bounding boxes, table structures, and key-value pairs), eliminating manual labeling and allows for precise error calculation and overall risk of each document based on the IDP oracle being tested.
Crucially, the generator is \emph{semantically deterministic} but \emph{visually stochastic}. While $z$ fixes the logical structure (the risk features), $\mathcal{G}$ introduces controlled variance in rendering (e.g., font rasterization, background texture). Thus, repeated evaluations of a fixed $z$ yield visually distinct instances, ensuring the search optimizes for robust structural failures rather than overfitting to specific pixel artifacts.

\subsection{System-Level Risk Oracle}
The IDP system under test is treated as a black-box oracle and comprises multiple pretrained models:
\emph{(i) OCR:} Text recognition and localization, \emph{(ii) Layout Analysis:} Document segmentation and block classification, \emph{(iii) Key–Value Extraction:} Semantic entity linking, \emph{(iv) Table Understanding:} Row/column structure recognition.

We define \textit{Risk} as the quantitative error between the system's output and the generator's ground truth. For a configuration $z$, the oracle computes a vector of $K$ normalized error components:
\[
\mathbf{r}(z) = [r_1(\mathcal{G}(z)), r_2(\mathcal{G}(z)), \dots, r_K(\mathcal{G}(z))] \in [0,1]^K
\]
Here, $K$ represents the total number of individual metrics monitored (e.g., OCR CER, Table IoU, Entity F1-score, etc.).
These components are aggregated by individual weights $w_k$ into a base risk score:
\[
R_{\text{base}}(z) = \sum_{k=1}^{K} w_k \cdot r_k(z)
\]

To account for the operational importance of structurally uncommon configurations, we define an optional rarity-based reweighting term that modulates the base risk score. Rarity is computed from available training statistics over selected risk features (e.g., density, noise level, pagination), reflecting how frequently a given feature interaction occurs in the configuration space.
When enabled, the final risk score takes the form:
\[
R(z) = R_{\text{base}}(z) \cdot \left(1 + \lambda \cdot \mathrm{Rarity}(z)\right)
\]
where $\lambda$ controls the influence of rarity. This mechanism allows the search objective to emphasize infrequent but potentially high-impact document structures, particular to the target domain.

\subsection{Failure Signatures}
To support diversity-oriented analysis, each evaluated configuration is mapped to a {Failure Signature}, capturing which categories of processing risk are activated by the interaction of risk features in $z$. Formally, let $\tau_k$ be a fixed decision threshold for the $k$-th error component. The failure signature is defined as:
\[
\mathrm{Signature}(z) =
\left[
\mathbb{I}(r_1(z) > \tau_1),
\dots,
\mathbb{I}(r_K(z) > \tau_K)
\right] \in \{0,1\}^K
\]
Each unique signature corresponds to a distinct failure type. The objective of search is to maximize the diversity of both the input risk features (via rarity) and the observed output failure signatures, while optimizing towards maximizing the final risk.

\subsection{Configuration Exclusivity}
From a search-based testing perspective, exclusivity measures the \emph{marginal fault discovery contribution} of a test generation strategy relative to competing strategies. Low exclusivity indicates generation of \emph{redundant test cases}, while high exclusivity reflects exploration of behaviorally distinct regions of the input space. 
To quantify how uniquely a method explores the document configuration space, we introduce an \textit{exclusivity percentage}: Given a method $A$ and the set of all other methods $\bar{A}$, at any budget step $t$, let $\mathcal{Z}_A(t)$ denote the set of configurations discovered by $A$ up to $t$, and $\mathcal{Z}_{\bar{A}}(t)$ the union of configurations discovered by all other methods. The exclusivity of $A$ is then,
\[
\mathrm{Excl}(A, t) =
\frac{|\mathcal{Z}_A(t) \setminus \mathcal{Z}_{\bar{A}}(t)|}
     {|\mathcal{Z}_A(t) \cup \mathcal{Z}_{\bar{A}}(t)|}.
\]

\subsection{Budgeted Search Protocol}
We evaluate all search strategies under a unified black-box interaction protocol (Algorithm~\ref{alg:rmd}) designed to reflect realistic validation constraints. Each strategy $\pi$ operates under a strict evaluation budget $B \ll |\mathcal{Z}|$, motivated by the high cost of executing production-grade IDP pipelines (e.g., GPU-intensive DL inference or paid OCR/LLM services). At each iteration, the strategy proposes a document configuration $z$ and receives feedback limited to a scalar risk score $R(z)$ and a discrete failure  $\mathrm{Signature}(z)$. No access is provided to internal oracle states, gradients, or intermediate outputs. As a result, search proceeds purely via observed input--output behavior, mirroring real-world testing scenarios involving remote, opaque systems. This standardized interface ensures fair comparison across methods, with observed performance differences arising solely from their exploration dynamics under budget constraints.

\begin{algorithm}[t]
\caption{Budgeted Risk Discovery}
\label{alg:rmd}
\begin{algorithmic}[1]
\Require Budget $B$, Generator $\mathcal{G}$, Oracle $IDP$, Search Policy $\pi$
\State Initialize archive $\mathcal{A} \gets \emptyset$
\For{$t = 1$ to $B$}
    \State Propose binary configuration $z_t \sim \pi(\mathcal{A})$
    \State \emph{Render \& Annotate:} $d_t, \text{GT}_t \sim \mathcal{G}(z_t)$
    \State \emph{Inference:} $\text{Pred}_t \gets IDP(d_t)$
    \State \emph{Compute Risk:} Calculate $\mathbf{r}(z_t)$ using $\text{GT}_t$ vs $\text{Pred}_t$
    \State Derive $R(z_t)$ and $\mathrm{Signature}(z_t)$
    \State Update archive $\mathcal{A} \gets \mathcal{A} \cup \{(z_t, R(z_t))\}$
    \State Update search policy $\pi$
\EndFor
\State \Return Archive $\mathcal{A}$
\end{algorithmic}
\end{algorithm}

\begin{table*}[t]
\centering
\small
\caption{Comprehensive risk discovery results under a fixed budget of 1000 oracle calls (3 seeds). T10 = top 10\% mean (mu) and std.dev (sd) scores; Rare = mean rarity score; UL = unique layouts discovered; AUC = normalized cumulative-max risk AUC; Ham = final Hamming diversity; Ent = final average feature entropy.}
\setlength{\tabcolsep}{2.2pt}
\renewcommand{\arraystretch}{1.}
\begin{tabular}{cc|cccccccccc|cccccccccc}
\hline
\textbf{Class} & \textbf{Method}
& \multicolumn{10}{c|}{\textbf{Single-Page (24D) Configuration}}
& \multicolumn{10}{c}{\textbf{Multi-Page (27D) Configuration}} \\
& 
& Max & Mean & Std & Rare & T10mu & T10sd & UL & AUC & Ham & Ent
& Max & Mean & Std & Rare & T10mu & T10sd & UL & AUC & Ham & Ent \\
\hline

\multirow{1}{*}{Random}
& Random
& 4.086 & 2.332 & 0.554 & 0.976 & 3.493 & \textbf{0.230} & 860 & 0.983 & \textbf{0.514 }& \textbf{1.513}
& 4.683 & 2.364 & 0.637 & 0.990 & 3.646 & 0.284 & 981 & 0.992 & 0.639 & 1.929 \\

\hline
\multirow{1}{*}{Local}
& SA
& 3.983 & 2.969 & 0.512 & 0.985 & 3.965 & 0.024 & 209 & 0.968 & 0.340 & 0.943
& 4.652 & 3.499 & 0.665 & 0.999 & 4.468 & 0.041 & 292 & 0.976 & 0.369 & 1.075 \\

\hline
\multirow{4}{*}{Evolution}
& GA-Explore
& \textbf{4.095} & 3.255 & 0.798 & \textbf{0.997} & 4.095 & 0.000 & 378 & 0.980 & 0.375 & 0.377
& 5.064 & 3.810 & 0.891 & 0.996 & 4.539 & 0.120 & 325 & 0.914 & 0.311 & 0.909 \\

& GA-Exploit
& 4.125 & \textbf{3.734} & 0.580 & 1.000 & 4.071 & 0.049 & 109 & 0.950 & 0.179 & 0.909
& 4.497 & 4.252 & 0.652 & 0.998 & 4.497 & 0.000 & 102 & 0.989 & 0.107 & 0.377 \\

& PSO
& 3.920 & 2.319 & 0.839 & 0.962 & 3.611 & 0.118 & 341 & 0.965 & 0.392 & 1.080
& 5.751 & 2.546 & 0.928 & 0.985 & 4.103 & 0.390 & 614 & 0.827 & 0.562 & 1.535 \\

& MAP-Elites
& 4.143 & 2.459 & 0.616 & 0.984 & 3.562 & 0.176 & 466 & 0.973 & 0.382 & 1.006
& 4.706 & 2.688 & 0.758 & 0.997 & 4.066 & 0.194 & 671 & 0.961 & 0.451 & 1.176 \\

\hline
\multirow{3}{*}{Bayesian}
& GP-EI
& 4.235 & 3.065 & 0.590 & 0.992 & 3.867 & 0.117 & \textbf{957 }& 0.969 & 0.499 & 1.422
& 5.323 & 3.410 & 0.607 & 0.991 & 4.333 & 0.284 & 994 & 0.966 & 0.560 & 1.680 \\

& GP-UCB
& 4.180 & 3.110 & 0.582 & 0.996 & 3.847 & 0.091 & \textbf{957 }& 0.979 & 0.493 & 1.404
& 5.146 & 3.426 & 0.599 & 0.990 & 4.326 & 0.223 & 993 & 0.948 & 0.538 & 1.623 \\

& TPE
& 4.303 & 3.475 & 0.621 & 0.992 & 3.956 & 0.035 & 187 & 0.953 & 0.224 & 0.687
& 4.437 & 3.936 & 0.721 & 0.996 & 4.437 & 0.000 & 242 & 0.972 & 0.240 & 0.793 \\

\hline
\multirow{3}{*}{RL}
& REINFORCE
& 4.179 & 2.326 & \textbf{0.851 }& 0.986 & 3.685 & 0.135 & 809 & 0.930 & 0.462 & 1.270
& 5.185 & 2.751 & 1.021 & 0.994 & 4.269 & 0.179 & 862 & 0.900 & 0.517 & 1.451 \\

& PPO-Risk
& 4.297 & 3.038 & 0.727 & 0.995 & 3.840 & 0.114 & 568 & 0.948 & 0.390 & 1.070
& 5.296 & 3.475 & 0.883 & 0.996 & 4.521 & 0.150 & 560 & 0.982 & 0.384 & 1.104 \\

& PPO-Div
& \textbf{4.395} & 2.874 & 0.769 & 0.990 & 3.850 & 0.151 & 687 & 0.933 & 0.420 & 1.157
& 4.921 & 3.215 & 0.891 & 0.992 & 4.410 & 0.161 & 724 & 0.984 & 0.446 & 1.263 \\

\hline
\multirow{2}{*}{Quantum}
& QAOA
& 4.131 & 2.662 & 0.757 & 0.996 & 3.843 & 0.110 & 651 & 0.974 & 0.425 & 1.214
& 4.909 & 3.122 & 0.839 & 0.998 & 4.439 & 0.152 & 701 & 0.963 & 0.482 & 1.380 \\

& QAOA-Corr
& 3.985 & 2.371 & 0.770 & 0.994 & 3.726 & 0.116 & 744 & \textbf{0.988 }& 0.462 & 1.297
& 4.758 & 2.816 & 0.796 & 0.996 & 4.222 & 0.188 & 761 & 0.957 & 0.503 & 1.443 \\

\hline
\end{tabular}
\label{tab:full_results}
\end{table*}

\section{Experiments}
We evaluate a diverse portfolio of search strategies for \emph{risk mode discovery} in IDP systems.  All methods operate under a strict and identical evaluation budget and interact with the system exclusively through the black-box protocol described in Section~\ref{section:methodology}. Our experimental design is guided by the following questions:
\begin{enumerate}
    \item How efficiently do different search strategies uncover high-risk and diverse failure modes under fixed budgets?
    \item Do different solvers exhibit complementary discovery behavior in terms of \emph{which} and \emph{when} failure modes are found?
    \item How does problem dimensionality (single vs.\ multi-page documents) affect solver behavior and relative performance?
\end{enumerate}


\subsection{Document Configuration}
The configuration space is instantiated using eight canonical document templates designed to reflect common structures encountered in financial and insurance workflows, including invoices, policy documents, statements, and correspondence.
Each template specifies an ordered sequence of semantic content blocks (e.g., \texttt{HEADER}, \texttt{KV}, \texttt{TABLE}, \texttt{TEXT}, \texttt{FOOTER}), capturing variations in layout density, block repetition, and content interleaving. This design enables controlled exploration of structural interactions such as text–table interleaving, repeated table regions, and sectioned layouts,while maintaining semantic realism across generated documents.

Two document configuration regimes are considered reflecting increasing structural complexity:
\subsubsection{Single-page:} The configuration TEMPLATE\_ID = 8, \\ NUM\_PAGES = 1, MAX\_KV = 3, MAX\_TEXT = 3, MAX\_TBL\_ROWS = 7, MAX\_TBL\_COLS = 7, MAX\_SUMMARY\_ROWS = 7, \\ MAX\_SUMMARY\_COLS = 7, (Counts), NOISE\_LEVEL = 3 (No / Low / Mid / High), TABLE\_CONTINUE\_PAGE = 1 (Yes / No), LAYOUT\_SPLIT = 3 (No / Soft / Hard), SUMMARY\_LAST\_PAGE = 1 (Yes/No) creates single page and lower content density, resulting in a 24-dim binary-encoded search space.

\subsubsection{Multi-page:} MAX\_PAGES=3, MAX\_KV = 7, MAX\_TEXT = 7 enables multi-page documents with higher content density and pagination effects with identical table and noise bounds, yielding a 27-dim binary configuration space. This regime introduces additional structural interactions such as table continuation and cross-page layout consistency.

 Due to compute increase on increasing dimension, KV and text content are made to repeat by a set value of 3 during synthetic document generation and annotation. To generate content,  \textit{Faker library} was employed, but it can be replaced with LLMs in future.

\begin{figure*}[t]
    \centering
    \includegraphics[width=0.95\linewidth]{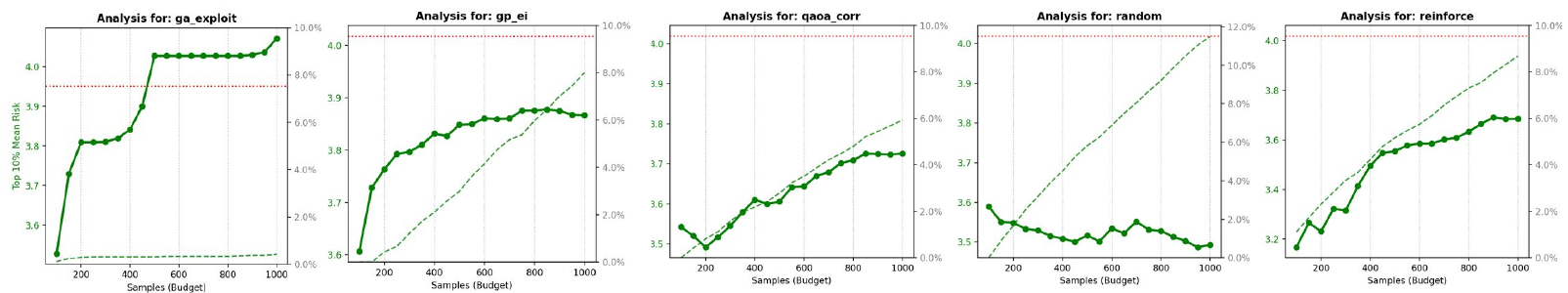}
    \vspace{-0.3cm}
    \caption{
    Risk-feature level exclusivity over budget for single-page (24D).
    Each curve reports the fraction exclusively discovered by a method relative to all others across same budget. The dotted red line represents the top 10\% mean risk over all other methods, while the dashed green line represents that method's top 10\% mean risk.
    }
    \label{fig:exclusivity_over_budget}
\end{figure*}

\subsection{Search Methods Evaluated}

\subsubsection{Random Baseline and Annealing:}
Random Search samples configurations uniformly from the design space and serves as a non-adaptive baseline. SA serves as a local-search baseline, using a geometric cooling schedule from $T_0=3.0$ to $T_{\text{end}}=0.05$. 

\subsubsection{Evolutionary and Swarm-Based Methods:}
Two GA variants are implemented using \texttt{DEAP} with population size 50. \emph{GA-Explore} employs high crossover ($p_c=0.9$), high mutation ($p_m=0.1$), and low selection pressure (tournament size 2), while \emph{GA-Exploit} uses lower crossover ($p_c=0.6$), low mutation ($p_m=0.01$), and stronger selection (tournament size 7).
Binary-PSO is implemented via \texttt{pyswarms} with inertia $w=0.7$ and cognitive/social coefficients $c_1=c_2=1.5$. In {MAP-Elites} implemented using \texttt{ribs} the archive is discretized into a $25\times25$ grid over selected structural descriptors, and new candidates are generated via Gaussian mutation ($\sigma=0.05$). 

\subsubsection{Bayesian Optimization:}
Surrogates are initialized with $n_{\text{init}}=100$ and sample topk candidates per iteration. We evaluate GP-EI and GP-UCB  using \texttt{scikit-learn}, and TPE is implemented in \texttt{Optuna}.

\subsubsection{Reinforcement Learning:}
RL agents are implemented using \texttt{Stable-Baselines3} and learn sampling policies directly over the binary configuration space.
{REINFORCE} maintains independent Bernoulli logits per bit and updates parameters via advantage-weighted policy gradients. Two PPO variants: {PPO-Risk} maximizes scalar risk, and {PPO-Div} augments the reward with an entropy coefficient 0.03 to encourage exploration.

\subsubsection{Quantum Optimization:}
QAOA is implemented using \textit{PennyLane} with the \texttt{lightning.kokkos} simulator under noiseless setting.
The quadratic surrogate in QUBO is learned iteratively via Factorization Machine (FM), which defines the cost Hamiltonian.
Two variational ansätze at fixed depth $p=2$: (i) {Standard mixer (QAOA)}, using the transverse-field mixer $\sum_i X_i$; and (ii) {Correlated mixer (QAOA-Corr)}, which augments the standard mixer with pairwise $X_iX_j$ interactions to induce structured, correlated bit flips. Both variants use COBYLA optimizer and measurement 5000 shots.

\subsection{Implementation Details}
All methods are evaluated under a fixed budget of $B=1000$ oracle calls per run. To ensure fairness, each experiment is repeated across 3 random seeds and results are averaged. All surrogate-based and learning-based methods use an initial random design of $n_{\text{init}}=100$ configurations. At each iteration, $50$ samples are selected by each method. (or is alternatively the population size for GA, PSO, and MAP-Elites). Repeated samples are retained for diversity metrics. This ensures comparable evaluation across methods per iteration. 

\subsection{Performance Summary}
Table~\ref{tab:full_results} summarizes the performance of all search strategies across single-page (24D) and multi-page (27D) configuration spaces. We evaluate methods along three complementary axes: (i) risk severity, (ii) budget efficiency, and (iii) structural diversity, reflecting the breadth of discovered failures under fixed evaluation budgets.
We report the following indicators:
\emph{Max Risk:} capturing the most severe failure discovered;
\emph{Top-10\% Risk (mean and stddev):}, characterizing the intensity and heterogeneity of the highest-risk regime;
\emph{Normalized AUC:} measuring how quickly high-risk configurations are discovered within the budget;
\emph{Unique Layouts:} serving as a proxy for structural coverage; and
\emph{Diversity Statistics:} (Final Hamming Diversity and Feature Entropy), quantifying dispersion in the binary configuration space.
\paragraph{4.4.1 Risk Severity and Exploitation Behavior:}
Results reveal a consistent separation between exploitative and exploratory strategies. Bayesian optimization (GP-EI, GP-UCB) and reward-driven RL (PPO-Risk) consistently achieve the highest \emph{Max Risk}, indicating effective exploitation once high-risk regions are located. This comes at the cost of diversity: low Top-10\% variance and limited structural coverage suggest repeated sampling of closely related configurations, with GA-Exploit exhibiting early collapse to a narrow set of failure modes. Exploration-oriented methods achieve lower mean risk but uncover a broader range of failures.
As dimensionality increases, policy-gradient methods and QAOA variants degrade more gracefully than local search (SA, PSO), indicating greater robustness to interaction-heavy and rugged risk landscapes.
\paragraph{4.4.2 Budget Efficiency and Discovery Timing:}
Normalized AUC reveals marked differences in discovery efficiency. BO, evolutionary, and QAOA achieve high AUC for single-page, indicating early localization of high-risk regions under strict budgets. In contrast, RL exhibits lower AUC which can be interpreted as requiring more samples. Similarly in multi-page, PSO despite eventually reaching highest risk has lower AUC, reflecting delayed convergence.
\paragraph{4.4.3 Structural Diversity and Coverage:}
Coverage-oriented metrics such as Unique Layout count, Hamming, and Average Entropy expose complementary solver behaviors. Random, GP and QAOA-corr achieve the broadest structural coverage, discovering a large fraction of unique layouts within the budget, whereas GA-Exploit, TPE, and SA scores indicate repeated sampling from a local optima.
\paragraph{4.4.4 Quantum Mixer Correlations:}
Comparing QAOA variants isolates the effect of interaction-aware mixing. Standard QAOA with a transverse-field mixer is outperformed by the Correlated mixer (QAOA-Corr), utilizing interaction-aware mixing to follow the topology of the risk landscape better. By encoding pairwise structure from QUBO, correlated mixing biases sampling toward jointly disruptive configurations, achieving a balance between early discovery, sustained diversity, and targeted exploration.

\subsection{Configuration-Level Discovery Dynamics}
\label{sec:config_discovery}
We now analyze discovery behavior \emph{directly in the configuration space} $\mathcal{Z}$. Specifically, we first study how different search strategies contribute \emph{novel structural configurations} over time, before downstream failure abstraction. 

\subsubsection{Exclusivity vs. Risk Concentration:} \label{sec:exclusivity} Figure~\ref{fig:exclusivity_over_budget} tracks configuration exclusivity over the evaluation budget in the 24D setting,. A clear trade-off emerges between exploration breadth and risk severity (Top-10\% Mean Risk) when comparing what was exclusively discovered by each method relative to all others. {Random Search} dominates exclusivity but suffers from weak risk concentration, producing unique but largely benign configurations. Conversely, exploitative methods like {GA} exhibits near-zero exclusivity despite high risk scores, indicating redundant rediscovery of identical failure modes found by other solvers. {GP-EI}, {REINFORCE}, and QAOA-Corr to some extent effectively navigate this Pareto frontier, maintaining distinct exploration paths (non-trivial exclusivity) while consistently locating high-severity failures. This balance between pure exploitation and random search is critical for maximizing information gain in budget-constrained validation.

\begin{figure*}[t]
    \centering
    \includegraphics[width=0.95\linewidth]{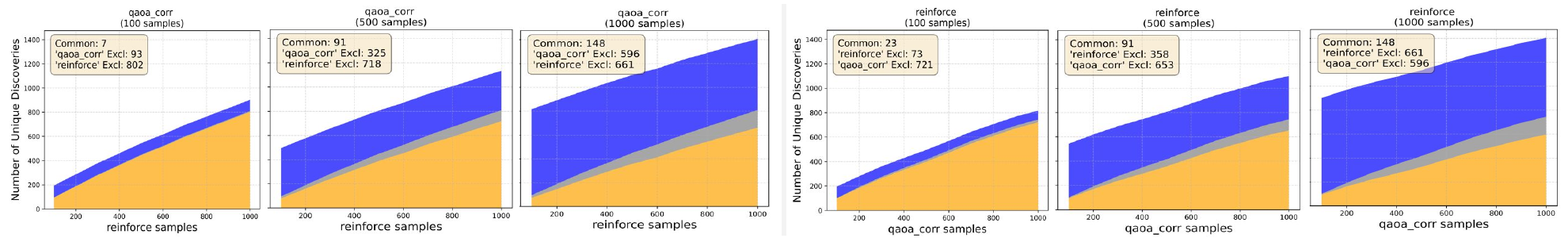}
    \vspace{-0.3cm}
    \caption{Cross-temporal risk features overlap between two methods.
    Each facet fixes one method at max budget, while the horizontal axis tracks the growth of discoveries made by the other method.
    Stacked Blue / Orange areas indicate features exclusive to each method, while Gray indicates those discovered in common.
    }
    \label{fig:cross_temporal_overlap}
\end{figure*}
\subsubsection{Cross-Temporal Dynamics:} \label{sec:cross_temporal}
Static metrics capture what is discovered, but fail to reveal when. To understand temporal efficiency, we analyze cross-temporal overlap, tracking how the discovery set of a fixed reference method, at snapshots of 100, 500, and 1000 samples is remains exclusive as the comparison Method evolves over time.
Figure~\ref{fig:cross_temporal_overlap} illustrates this interaction between QAOA-Corr and REINFORCE as an example. We observe the following characteristics: \emph{(i) Persistent Exclusivity:} When QAOA-Corr is fixed at just 100 samples (left panel), REINFORCE rediscovers only 7 of those configurations even after 1000 evaluations, leaving 93 exclusive to the early quantum search. Similarly when REINFORCE is fixed at 100 (right panel), QAOA-Corr rediscovers only 23 of those configurations even after 1000 evaluations, leaving 73 exclusive to the early quantum search. The exclusivity numbers increase with more samples discovered by either method incrementally. \emph{(ii Asymmetric Convergence:} The dynamic is non-reciprocal. At the 1000-sample mark, REINFORCE maintains 661 unique discoveries that QAOA-Corr has not found, while QAOA-Corr retains 596 exclusive configurations. \emph{(iii) Slow Saturation:} The "Common" region (gray) grows linearly rather than exponentially, indicating that neither method rapidly invalidates the other. Even at high budgets, the majority of the discovery space remains partitioned (Blue vs. Orange regions), confirming that these solvers traverse structurally distinct trajectories throughout the search. This is found to generalize across all solver pair comparisons.
This temporal persistence proves that a portfolio is not just about final coverage, but about time-to-value: a method can secure certain unique failures early in the budget that the other may effectively never find, and vice-versa.

\subsection{Risk Modes and Failure Characterization}

To analyze distinct failure mechanisms beyond $z$, we define a \textit{Risk Mode} as a semantic conjunction / interaction of input configuration descriptors (e.g., layout topology, density regimes) and observable failure outcomes. Unlike low-dimensional failure signatures, risk modes map valid system failures back to their structural causes. While multiple configurations may share the same failure signature, they correspond to distinct risk modes reflecting different  mechanisms. 

\subsubsection{Risk Mode Discovery Statistics:} Across all methods,
it is observed dominant modes reflect systematic stress points in dense, table-heavy layouts rather than isolated anomalies. The most frequently discovered risk modes are dominated by table truncation in high-density, high-noise regimes across all templates. Specifically, TEMPLATE\_ID:1 combined with LAYOUT:HARD\_SPLIT constitutes the single largest failure cluster (1,187 instances), indicating a systematic system vulnerability to aggressive segmentation.
Comparing solvers, we find that those explicitly designed to benefit from entropy or correlation achieve the broadest coverage. {REINFORCE} identifies the largest fraction (69.0\%), followed by {QAOA-Corr} (58.2\%) and {PPO-Div} (56.5\%). Purely exploitative optimizers (GA-Exploit, SA) discover fewer than 30\% of known modes, converging to a narrow set of easy-to-find failures. Overall a total of {368 unique risk modes} were identified, reflecting a rich and heterogeneous failure landscape.

\begin{table*}[t]
\centering
\small
\setlength{\tabcolsep}{2pt}
\renewcommand{\arraystretch}{1.}
\caption{Pairwise configuration-level overlap of discovered core risk modes for 24D single-page. For each method pair, we report the number of shared configurations and the respective exclusive discoveries of the row and column methods (Shared/Row/Col). Cell color and intensity encode dominance and asymmetry in discovery, providing a compact view of relative exploratory bias and coverage behavior under identical budgets. To read col-wise: Green if Col $>$ Row, Red if Row $>$ Col, Orange if Col $=$ Row.}
\label{tab:pairwise_exclusive_modes_full}
\begin{tabular}{l|llllllllllllll}
\hline
\textbf{Method} & \textbf{GA-Explt} & \textbf{GA-Explr} & \textbf{GP-EI} & \textbf{GP-UCB} & \textbf{MAP-El} & \textbf{PPO-Div} & \textbf{PPO-Risk} & \textbf{PSO} & \textbf{QAOA} & \textbf{QAOA-C} & \textbf{Random} & \textbf{REIN} & \textbf{SA} & \textbf{TPE} \\
\hline
\textbf{GA-Exploit} & -- & 
\gradcell{20}{0}{3}{7} & \gradcell{18}{2}{6}{7} & \gradcell{18}{2}{6}{7} & \gradcell{17}{3}{2}{7} & \gradcell{20}{0}{4}{7} & \gradcell{20}{0}{4}{7} & \gradcell{19}{1}{4}{7} & \gradcell{18}{2}{7}{7} & \gradcell{19}{1}{7}{7} & \gradcell{18}{2}{6}{7} & \gradcell{20}{0}{4}{7} & \gradcell{16}{4}{5}{7} & \gradcell{16}{4}{6}{7} \\

\textbf{GA-Explore} & \gradcell{20}{3}{0}{7} & -- & 
\gradcell{21}{2}{3}{7} & \gradcell{21}{2}{3}{7} & \gradcell{19}{4}{0}{7} & \gradcell{23}{0}{1}{7} & \gradcell{23}{0}{1}{7} & \gradcell{22}{1}{1}{7} & \gradcell{21}{2}{4}{7} & \gradcell{22}{1}{4}{7} & \gradcell{21}{2}{3}{7} & \gradcell{23}{0}{1}{7} & \gradcell{18}{5}{3}{7} & \gradcell{19}{4}{3}{7} \\

\textbf{GP-EI} & \gradcell{18}{6}{2}{7} & \gradcell{21}{3}{2}{7} & -- & 
\gradcell{24}{0}{0}{7} & \gradcell{18}{6}{1}{7} & \gradcell{21}{3}{3}{7} & \gradcell{21}{3}{3}{7} & \gradcell{20}{4}{3}{7} & \gradcell{24}{0}{1}{7} & \gradcell{24}{0}{2}{7} & \gradcell{24}{0}{0}{7} & \gradcell{21}{3}{3}{7} & \gradcell{21}{3}{0}{7} & \gradcell{22}{2}{0}{7} \\

\textbf{GP-UCB} & \gradcell{18}{6}{2}{7} & \gradcell{21}{3}{2}{7} & \gradcell{24}{0}{0}{7} & -- & 
\gradcell{18}{6}{1}{7} & \gradcell{21}{3}{3}{7} & \gradcell{21}{3}{3}{7} & \gradcell{20}{4}{3}{7} & \gradcell{24}{0}{1}{7} & \gradcell{24}{0}{2}{7} & \gradcell{24}{0}{0}{7} & \gradcell{21}{3}{3}{7} & \gradcell{21}{3}{0}{7} & \gradcell{22}{2}{0}{7} \\

\textbf{MAP-Elite} & \gradcell{17}{2}{3}{7} & \gradcell{19}{0}{4}{7} & \gradcell{18}{1}{6}{7} & \gradcell{18}{1}{6}{7} & -- & 
\gradcell{19}{0}{5}{7} & \gradcell{19}{0}{5}{7} & \gradcell{18}{1}{5}{7} & \gradcell{18}{1}{7}{7} & \gradcell{19}{0}{7}{7} & \gradcell{18}{1}{6}{7} & \gradcell{19}{0}{5}{7} & \gradcell{15}{4}{6}{7} & \gradcell{16}{3}{6}{7} \\

\textbf{PPO-Div} & \gradcell{20}{4}{0}{7} & \gradcell{23}{1}{0}{7} & \gradcell{21}{3}{3}{7} & \gradcell{21}{3}{3}{7} & \gradcell{19}{5}{0}{7} & -- & 
\gradcell{24}{0}{0}{7} & \gradcell{23}{1}{0}{7} & \gradcell{22}{2}{3}{7} & \gradcell{23}{1}{3}{7} & \gradcell{21}{3}{3}{7} & \gradcell{24}{0}{0}{7} & \gradcell{18}{6}{3}{7} & \gradcell{19}{5}{3}{7} \\

\textbf{PPO-Risk} & \gradcell{20}{4}{0}{7} & \gradcell{23}{1}{0}{7} & \gradcell{21}{3}{3}{7} & \gradcell{21}{3}{3}{7} & \gradcell{19}{5}{0}{7} & \gradcell{24}{0}{0}{7} & -- & 
\gradcell{23}{1}{0}{7} & \gradcell{22}{2}{3}{7} & \gradcell{23}{1}{3}{7} & \gradcell{21}{3}{3}{7} & \gradcell{24}{0}{0}{7} & \gradcell{18}{6}{3}{7} & \gradcell{19}{5}{3}{7} \\

\textbf{PSO} & \gradcell{19}{4}{1}{7} & \gradcell{22}{1}{1}{7} & \gradcell{20}{3}{4}{7} & \gradcell{20}{3}{4}{7} & \gradcell{18}{5}{1}{7} & \gradcell{23}{0}{1}{7} & \gradcell{23}{0}{1}{7} & -- & 
\gradcell{21}{2}{4}{7} & \gradcell{22}{1}{4}{7} & \gradcell{20}{3}{4}{7} & \gradcell{23}{0}{1}{7} & \gradcell{18}{5}{3}{7} & \gradcell{19}{4}{3}{7} \\

\textbf{QAOA} & \gradcell{18}{7}{2}{7} & \gradcell{21}{4}{2}{7} & \gradcell{24}{1}{0}{7} & \gradcell{24}{1}{0}{7} & \gradcell{18}{7}{1}{7} & \gradcell{22}{3}{2}{7} & \gradcell{22}{3}{2}{7} & \gradcell{21}{4}{2}{7} & -- & 
\gradcell{25}{0}{1}{7} & \gradcell{24}{1}{0}{7} & \gradcell{22}{3}{2}{7} & \gradcell{21}{4}{0}{7} & \gradcell{22}{3}{0}{7} \\

\textbf{QAOA-Corr} & \gradcell{19}{7}{1}{7} & \gradcell{22}{4}{1}{7} & \gradcell{24}{2}{0}{7} & \gradcell{24}{2}{0}{7} & \gradcell{19}{7}{0}{7} & \gradcell{23}{3}{1}{7} & \gradcell{23}{3}{1}{7} & \gradcell{22}{4}{1}{7} & \gradcell{25}{1}{0}{7} & -- & 
\gradcell{24}{2}{0}{7} & \gradcell{23}{3}{1}{7} & \gradcell{21}{5}{0}{7} & \gradcell{22}{4}{0}{7} \\

\textbf{Random} & \gradcell{18}{6}{2}{7} & \gradcell{21}{3}{2}{7} & \gradcell{24}{0}{0}{7} & \gradcell{24}{0}{0}{7} & \gradcell{18}{6}{1}{7} & \gradcell{21}{3}{3}{7} & \gradcell{21}{3}{3}{7} & \gradcell{20}{4}{3}{7} & \gradcell{24}{0}{1}{7} & \gradcell{24}{0}{2}{7} & -- & 
\gradcell{21}{3}{3}{7} & \gradcell{21}{3}{0}{7} & \gradcell{22}{2}{0}{7} \\

\textbf{REINFORCE} & \gradcell{20}{4}{0}{7} & \gradcell{23}{1}{0}{7} & \gradcell{21}{3}{3}{7} & \gradcell{21}{3}{3}{7} & \gradcell{19}{5}{0}{7} & \gradcell{24}{0}{0}{7} & \gradcell{24}{0}{0}{7} & \gradcell{23}{1}{0}{7} & \gradcell{22}{2}{3}{7} & \gradcell{23}{1}{3}{7} & \gradcell{21}{3}{3}{7} & -- & 
\gradcell{18}{6}{3}{7} & \gradcell{19}{5}{3}{7} \\

\textbf{SA} & \gradcell{16}{5}{4}{7} & \gradcell{18}{3}{5}{7} & \gradcell{21}{0}{3}{7} & \gradcell{21}{0}{3}{7} & \gradcell{15}{6}{4}{7} & \gradcell{18}{3}{6}{7} & \gradcell{18}{3}{6}{7} & \gradcell{18}{3}{5}{7} & \gradcell{21}{0}{4}{7} & \gradcell{21}{0}{5}{7} & \gradcell{21}{0}{3}{7} & \gradcell{18}{3}{6}{7} & -- & 
\gradcell{21}{0}{1}{7} \\

\textbf{TPE} & \gradcell{16}{6}{4}{7} & \gradcell{19}{3}{4}{7} & \gradcell{22}{0}{2}{7} & \gradcell{22}{0}{2}{7} & \gradcell{16}{6}{3}{7} & \gradcell{19}{3}{5}{7} & \gradcell{19}{3}{5}{7} & \gradcell{19}{3}{4}{7} & \gradcell{22}{0}{3}{7} & \gradcell{22}{0}{4}{7} & \gradcell{22}{0}{2}{7} & \gradcell{19}{3}{5}{7} & \gradcell{21}{1}{0}{7} & -- \\
\hline
\end{tabular}
\end{table*}

\subsubsection{Exclusive Risk Modes:}
\label{sec:exclusive_modes}
We distinguish between unique configurations and \emph{core exclusivity} (unique semantic failure classes). At the level of fine-grained risk modes, all methods appear to discover exclusive failures; however, the modes differ only by minor predicate refinements and are subsets of semantically equivalent modes. To avoid over-counting such variations, we introduce the notion of \emph{core risk modes}, where each risk mode is flattened to its essential predicates—failure type, density regime, noise regime, and layout interaction, while discarding incidental configuration details. This abstraction groups structurally equivalent failures while preserving the dominant causal factors. Under this definition, the total number of unique core risk modes reduces to 27. 

While no single method isolates a core mode entirely, pairwise comparisons in Table~\ref{tab:pairwise_exclusive_modes_full} reveal interesting observations: \textit{Quantum Utility:} In the 24D configuration, {QAOA-Corr} emerges as the best explorer, achieving a positive \textit{win rate} against other methods. For e.g., it identifies unique failure modes missed by {REINFORCE} (3 unique by QAOA-Corr vs. 1 by REINFORCE) and {SA} (5 unique vs. 0), suggesting that the method design mechanics can play an important role in access to different parts of the risk landscape. These results position QAOA as a possible additive component for validation portfolios, capable of patrolling structural boundaries that standard baselines may miss out.
\textit{Exploitative Collapse:} Purely exploitative methods ({GA-Exploit, SA, TPE}) act as strict subsets of exploratory strategies,
empirically showing that local search heuristics fail to break out of the dominant, easy-to-find failure basins. \textit{Algorithmic Equivalence:} In this domain, certain distinct algorithms converge to identical semantic coverage. {GP-EI} and {Random Search} exhibit perfect overlap (0 unique differences), implying that Gaussian Process variance estimation offers little advantage over uniform sampling for core mode discovery here (note GP still outperforms Random in max and mean risk as seen in Table \ref{tab:full_results}).

\subsection{Predictive Modeling for Future Risk}
To assess risk landscape learnability and sample quality, we trained Random Forest regressors using 70\% train ($N=700$) generated by each strategy, while collating 30\% from each to form the hold-out validation set. Table~\ref{tab:rf_model_performance} summarizes the results. The regressor trained on the full aggregate dataset (Upper Bound) achieves an $R^2$ of \textit{0.915}. This confirms that IDP risk are deterministic outcomes of structural feature interactions that can be robustly modeled given sufficient data. Methods prioritizing structural diversity (PPO-Div, QAOA-Corr) generate significantly better training data ($R^2 \approx 0.79$) than purely exploitative strategies like SA ($R^2 = 0.62$). This indicates that broader coverage captures global risk dependencies more effectively than narrow optimization paths. Crucially, portfolio = 0.832 {(samples from 101-300 from all 14 methods, leaving first 100 as init warmup)} model outperforms the best single-method (PPO-Div = 0.795) even when strictly controlling for training size ($N=700$), 
showing that solver complementarity is instantaneous: the aggregated initial trajectories of a diverse portfolio provide a higher-fidelity signal for risk modeling than the fully converged set of any single heuristic.
\begin{table}[h!]\centering\small\setlength{\tabcolsep}{7pt}\renewcommand{\arraystretch}{1.1}\caption{Predictive performance of RF Regressors. To ensure fair comparison, all strategy-specific models and Combined subsets use exactly 700 training samples. Portfolio (200 samples per method) across all methods yields superior generalization with just early search phase aggregation.} \label{tab:rf_model_performance}\begin{tabular}{llccc}\hline\textbf{Category} & \textbf{Method} & $\mathbf{R^2}$ & \textbf{MAE} & \textbf{RMSE} \\\textit{Upper Bound} & \textbf{(Full, N=9800)} & \textbf{0.915} & \textbf{0.159} & \textbf{0.245} \\ \textit{Portfolio} & \textbf{(Random, N=700)} & \textbf{0.852} & \textbf{0.225} & \textbf{0.323} \\ \textit{Portfolio} & \textbf{(101-300, N=700)} & \textbf{0.832} & \textbf{0.257} & \textbf{0.344} \\ \textit{RL (Diversity)} & PPO-Div & 0.795 & 0.285 & 0.380 \\ \textit{Quantum} & QAOA-Corr & 0.792 & 0.295 & 0.383 \\ \textit{Swarm} & PSO & 0.722 & 0.338 & 0.442 \\ \textit{Evolutionary} & GA-Explore & 0.715 & 0.336 & 0.448 \\ \textit{QD} & MAP-Elites & 0.714 & 0.330 & 0.449 \\ \textit{Baseline} & Random Search & 0.680 & 0.370 & 0.475 \\ \textit{Bayesian Opt} & GP-EI & 0.664 & 0.383 & 0.486 \\ \textit{Local Search} & SA & 0.624 & 0.382 & 0.515 \\\hline\end{tabular}\end{table}

\vspace{-0.2cm}
\section{Discussion and Conclusion}
This work reformulates Intelligent Document Processing (IDP) validation as a budgeted, structure-aware combinatorial search problem. By advancing SBST through diversity-driven exploration, we demonstrate that a singular focus on peak-risk metrics obscures critical robustness gaps. We now discuss the implications of these findings for IDP validation and robust system design:

\emph{(i) SBST for IDP} mandates a shift from peak-risk maximization to behavioral diversity. Unlike simple function optimization, IDP failures arise from complex interactions between layout, noise, and content. Our results show that while exploitative heuristics efficiently maximize scalar error, they often collapse into narrow failure basins. Robust IDP validation therefore requires maximizing the diversity of discovered mechanisms, prioritizing the number of unique and exclusive failure types over the severity of any single instance. These dynamics are governed by the strict budget constraints ($B \ll |\mathcal{Z}|$) of black-box validation. For the practical testing of costly pipelines where exhaustive search is impossible, the efficiency of diverse discovery is the governing metric.

\emph{(ii) Solver Portfolio enables closed-loop system hardening:} We identified a functional dichotomy: exploitative methods efficiently sweep core risks, while explorers are strictly necessary to uncover the \textit{long tail} of idiosyncratic failures. Robust validation requires layering cost-effective heuristics to patrol known weaknesses with diversity-driven solvers to illuminate structural blind spots.

\emph{(iii) Moving from reactive to proactive robustness:} By leveraging the discovered failure manifold, practitioners can synthetically generate adversarial-yet-realistic configurations to target predicted weak zones. Possible extensions are fine-tuning in continuous cycles from development to deployment, and transforming validation to a diagnostic robustness engine that closes reliability gaps without awaiting real-world incidents.

\emph{(iv) Quantum-Enhanced Exploration:} Within this portfolio, QAOA-Corr provides a distinct advantage via structure-aware quantum interference. Unlike local search which relies on stepwise bit-flips, the correlated mixer encodes learned dependencies to \textit{tunnel} through invalid configuration barriers, locating deep structural failure pockets that remain topologically inaccessible to standard heuristics.

\emph{(v) Landscape Determinism} The high predictive accuracy of the RFRegressor ($R^2 > 0.91$) confirms that IDP failures are not stochastic but governed by learnable feature interactions. Diverse solvers traverse distinct paths yet converge on valid failure modes shows the risk landscape is structured and predictable.

We envision extending this work by
integrating LLMs as semantic oracles to assess higher-order reasoning failures beyond structural rule violations and improve quantum optimization by learning adaptive mixers to systematically isolate how correlations enhance exploratory coverage.

\bibliographystyle{ACM-Reference-Format}
\bibliography{main}


\begin{thebibliography}{44}


\ifx \showCODEN    \undefined \def \showCODEN     #1{\unskip}     \fi
\ifx \showISBNx    \undefined \def \showISBNx     #1{\unskip}     \fi
\ifx \showISBNxiii \undefined \def \showISBNxiii  #1{\unskip}     \fi
\ifx \showISSN     \undefined \def \showISSN      #1{\unskip}     \fi
\ifx \showLCCN     \undefined \def \showLCCN      #1{\unskip}     \fi
\ifx \shownote     \undefined \def \shownote      #1{#1}          \fi
\ifx \showarticletitle \undefined \def \showarticletitle #1{#1}   \fi
\ifx \showURL      \undefined \def \showURL       {\relax}        \fi
\providecommand\bibfield[2]{#2}
\providecommand\bibinfo[2]{#2}
\providecommand\natexlab[1]{#1}
\providecommand\showeprint[2][]{arXiv:#2}

\bibitem[Afzal et~al\mbox{.}(2009)]%
        {afzal2009systematic}
\bibfield{author}{\bibinfo{person}{Wasif Afzal}, \bibinfo{person}{Richard Torkar}, {and} \bibinfo{person}{Robert Feldt}.} \bibinfo{year}{2009}\natexlab{}.
\newblock \showarticletitle{A systematic review of search-based testing for non-functional system properties}.
\newblock \bibinfo{journal}{\emph{Information and Software Technology}} \bibinfo{volume}{51}, \bibinfo{number}{6} (\bibinfo{year}{2009}), \bibinfo{pages}{957--976}.
\newblock


\bibitem[Alexiou et~al\mbox{.}(2023)]%
        {alexiou2023evaluation}
\bibfield{author}{\bibinfo{person}{Michail~S Alexiou}, \bibinfo{person}{Euripides~GM Petrakis}, {and} \bibinfo{person}{Nikolaos~G Bourbakis}.} \bibinfo{year}{2023}\natexlab{}.
\newblock \showarticletitle{An evaluation of table detection methods in document images}. In \bibinfo{booktitle}{\emph{2023 IEEE 35th International Conference on Tools with Artificial Intelligence (ICTAI)}}. IEEE, \bibinfo{pages}{54--63}.
\newblock


\bibitem[Bergstra et~al\mbox{.}(2011)]%
        {bergstra2011tpe}
\bibfield{author}{\bibinfo{person}{James Bergstra}, \bibinfo{person}{R{\'e}mi Bardenet}, \bibinfo{person}{Yoshua Bengio}, {and} \bibinfo{person}{Bal{\'a}zs K{\'e}gl}.} \bibinfo{year}{2011}\natexlab{}.
\newblock \showarticletitle{Algorithms for Hyper-Parameter Optimization}. In \bibinfo{booktitle}{\emph{NeurIPS}}.
\newblock


\bibitem[Biswas et~al\mbox{.}(2021)]%
        {10.1007/978-3-030-86334-0_36}
\bibfield{author}{\bibinfo{person}{Sanket Biswas}, \bibinfo{person}{Pau Riba}, \bibinfo{person}{Josep Llad\'{o}s}, {and} \bibinfo{person}{Umapada Pal}.} \bibinfo{year}{2021}\natexlab{}.
\newblock \showarticletitle{DocSynth: A Layout Guided Approach for Controllable Document Image Synthesis}. In \bibinfo{booktitle}{\emph{Document Analysis and Recognition – ICDAR 2021: 16th International Conference, Lausanne, Switzerland, September 5–10, 2021, Proceedings, Part III}}. \bibinfo{publisher}{Springer-Verlag}, \bibinfo{address}{Berlin, Heidelberg}.
\newblock
\showISBNx{978-3-030-86333-3}
\href{https://doi.org/10.1007/978-3-030-86334-0_36}{doi:\nolinkurl{10.1007/978-3-030-86334-0_36}}


\bibitem[Chen et~al\mbox{.}(2024)]%
        {chen2024rodla}
\bibfield{author}{\bibinfo{person}{Yufan Chen}, \bibinfo{person}{Jiaming Zhang}, \bibinfo{person}{Kunyu Peng}, \bibinfo{person}{Junwei Zheng}, \bibinfo{person}{Ruiping Liu}, \bibinfo{person}{Philip Torr}, {and} \bibinfo{person}{Rainer Stiefelhagen}.} \bibinfo{year}{2024}\natexlab{}.
\newblock \showarticletitle{Rodla: Benchmarking the robustness of document layout analysis models}. In \bibinfo{booktitle}{\emph{Proceedings of the IEEE/CVF Conference on Computer Vision and Pattern Recognition}}. \bibinfo{pages}{15556--15566}.
\newblock


\bibitem[Cui et~al\mbox{.}(2021)]%
        {cui2021document}
\bibfield{author}{\bibinfo{person}{Lei Cui}, \bibinfo{person}{Yiheng Xu}, \bibinfo{person}{Tengchao Lv}, {and} \bibinfo{person}{Furu Wei}.} \bibinfo{year}{2021}\natexlab{}.
\newblock \showarticletitle{Document AI: Benchmarks, Models and Applications}.
\newblock \bibinfo{journal}{\emph{arXiv preprint arXiv:2111.08609}} (\bibinfo{year}{2021}).
\newblock


\bibitem[Deb et~al\mbox{.}(2002)]%
        {deb2002fast}
\bibfield{author}{\bibinfo{person}{Kalyanmoy Deb}, \bibinfo{person}{Amrit Pratap}, \bibinfo{person}{Sameer Agarwal}, {and} \bibinfo{person}{TAMT Meyarivan}.} \bibinfo{year}{2002}\natexlab{}.
\newblock \showarticletitle{A fast and elitist multiobjective genetic algorithm: NSGA-II}.
\newblock \bibinfo{journal}{\emph{IEEE transactions on evolutionary computation}} \bibinfo{volume}{6}, \bibinfo{number}{2} (\bibinfo{year}{2002}), \bibinfo{pages}{182--197}.
\newblock


\bibitem[Deutsch(1985)]%
        {deutsch1985quantum}
\bibfield{author}{\bibinfo{person}{David Deutsch}.} \bibinfo{year}{1985}\natexlab{}.
\newblock \showarticletitle{Quantum theory, the Church--Turing principle and the universal quantum computer}.
\newblock \bibinfo{journal}{\emph{Proceedings of the Royal Society of London. A. Mathematical and Physical Sciences}} \bibinfo{volume}{400}, \bibinfo{number}{1818} (\bibinfo{year}{1985}), \bibinfo{pages}{97--117}.
\newblock


\bibitem[Farhi et~al\mbox{.}(2014)]%
        {farhi2014qaoa}
\bibfield{author}{\bibinfo{person}{Edward Farhi}, \bibinfo{person}{Jeffrey Goldstone}, {and} \bibinfo{person}{Sam Gutmann}.} \bibinfo{year}{2014}\natexlab{}.
\newblock \showarticletitle{A Quantum Approximate Optimization Algorithm}.
\newblock \bibinfo{journal}{\emph{arXiv preprint arXiv:1411.4028}} (\bibinfo{year}{2014}).
\newblock


\bibitem[Fatehi et~al\mbox{.}(2023)]%
        {fatehi2023towards}
\bibfield{author}{\bibinfo{person}{M. Fatehi} {et~al\mbox{.}}} \bibinfo{year}{2023}\natexlab{}.
\newblock \showarticletitle{Towards Adversarial Attacks for Clinical Document Classification}.
\newblock \bibinfo{journal}{\emph{Electronics}} \bibinfo{volume}{12}, \bibinfo{number}{1} (\bibinfo{year}{2023}).
\newblock


\bibitem[Gilani et~al\mbox{.}(2017)]%
        {gilani2017table}
\bibfield{author}{\bibinfo{person}{Azka Gilani}, \bibinfo{person}{Shah~Rukh Qasim}, \bibinfo{person}{Imran Malik}, {and} \bibinfo{person}{Faisal Shafait}.} \bibinfo{year}{2017}\natexlab{}.
\newblock \showarticletitle{Table detection using deep learning}. In \bibinfo{booktitle}{\emph{2017 14th IAPR international conference on document analysis and recognition (ICDAR)}}, Vol.~\bibinfo{volume}{1}. IEEE, \bibinfo{pages}{771--776}.
\newblock


\bibitem[Glover and Kochenberger(2018)]%
        {glover2018qubo}
\bibfield{author}{\bibinfo{person}{Fred Glover} {and} \bibinfo{person}{Gary Kochenberger}.} \bibinfo{year}{2018}\natexlab{}.
\newblock \showarticletitle{A Tutorial on Formulating and Using {QUBO} Models}.
\newblock \bibinfo{journal}{\emph{arXiv preprint arXiv:1811.11538}} (\bibinfo{year}{2018}).
\newblock


\bibitem[Harikrishnan et~al\mbox{.}({[n.\,d.]})]%
        {harikrishnan2025docgenie}
\bibfield{author}{\bibinfo{person}{PM Harikrishnan}, \bibinfo{person}{Siddartha Reddy}, \bibinfo{person}{Goutham Vignesh}, \bibinfo{person}{Rohit Agrawal}, {and} \bibinfo{person}{Vishal Vaddina}.} \bibinfo{year}{[n.\,d.]}\natexlab{}.
\newblock \showarticletitle{DocGenie: A Framework for High-Fidelity Synthetic Document Generation via Seed-Guided Multimodal LLM and Document-Aware Evaluation}. In \bibinfo{booktitle}{\emph{Synthetic Data for Computer Vision Workshop@ CVPR 2025}}.
\newblock


\bibitem[Harman and Jones(2001)]%
        {harman2001sbse}
\bibfield{author}{\bibinfo{person}{Mark Harman} {and} \bibinfo{person}{Bryan~F. Jones}.} \bibinfo{year}{2001}\natexlab{}.
\newblock \showarticletitle{Search-based software engineering}.
\newblock \bibinfo{journal}{\emph{Information and Software Technology}} \bibinfo{volume}{43}, \bibinfo{number}{14} (\bibinfo{year}{2001}), \bibinfo{pages}{833--839}.
\newblock
\href{https://doi.org/10.1016/S0950-5849(01)00189-6}{doi:\nolinkurl{10.1016/S0950-5849(01)00189-6}}


\bibitem[Holland(1975)]%
        {holland1975adaptation}
\bibfield{author}{\bibinfo{person}{John~H. Holland}.} \bibinfo{year}{1975}\natexlab{}.
\newblock \bibinfo{booktitle}{\emph{Adaptation in Natural and Artificial Systems}}.
\newblock \bibinfo{publisher}{University of Michigan Press}.
\newblock


\bibitem[Huang et~al\mbox{.}(2022)]%
        {huang2022layoutlmv3}
\bibfield{author}{\bibinfo{person}{Yupan Huang}, \bibinfo{person}{Tengchao Lv}, \bibinfo{person}{Lei Cui}, \bibinfo{person}{Yutong Lu}, {and} \bibinfo{person}{Furu Wei}.} \bibinfo{year}{2022}\natexlab{}.
\newblock \showarticletitle{Layoutlmv3: Pre-training for document ai with unified text and image masking}. In \bibinfo{booktitle}{\emph{Proceedings of the 30th ACM international conference on multimedia}}. \bibinfo{pages}{4083--4091}.
\newblock


\bibitem[Huang et~al\mbox{.}(2019)]%
        {huang2019icdar2019}
\bibfield{author}{\bibinfo{person}{Zheng Huang}, \bibinfo{person}{Kai Chen}, \bibinfo{person}{Jianhua He}, \bibinfo{person}{Xiang Bai}, \bibinfo{person}{Dimosthenis Karatzas}, \bibinfo{person}{Shijian Lu}, {and} \bibinfo{person}{CV Jawahar}.} \bibinfo{year}{2019}\natexlab{}.
\newblock \showarticletitle{Icdar2019 competition on scanned receipt ocr and information extraction}. In \bibinfo{booktitle}{\emph{2019 International Conference on Document Analysis and Recognition (ICDAR)}}. IEEE, \bibinfo{pages}{1516--1520}.
\newblock


\bibitem[Jaume et~al\mbox{.}(2019)]%
        {jaume2019funsd}
\bibfield{author}{\bibinfo{person}{Guillaume Jaume}, \bibinfo{person}{Hazim~Kemal Ekenel}, {and} \bibinfo{person}{Jean-Philippe Thiran}.} \bibinfo{year}{2019}\natexlab{}.
\newblock \showarticletitle{Funsd: A dataset for form understanding in noisy scanned documents}. In \bibinfo{booktitle}{\emph{2019 International Conference on Document Analysis and Recognition Workshops (ICDARW)}}, Vol.~\bibinfo{volume}{2}. IEEE, \bibinfo{pages}{1--6}.
\newblock


\bibitem[Jones et~al\mbox{.}(1998)]%
        {jones1998ego}
\bibfield{author}{\bibinfo{person}{Donald~R. Jones}, \bibinfo{person}{Matthias Schonlau}, {and} \bibinfo{person}{William~J. Welch}.} \bibinfo{year}{1998}\natexlab{}.
\newblock \showarticletitle{Efficient Global Optimization of Expensive Black-Box Functions}.
\newblock \bibinfo{journal}{\emph{Journal of Global Optimization}}  \bibinfo{volume}{13} (\bibinfo{year}{1998}), \bibinfo{pages}{455--492}.
\newblock


\bibitem[Kadowaki and Nishimori(1998)]%
        {kadowaki1998quantumannealing}
\bibfield{author}{\bibinfo{person}{Tadashi Kadowaki} {and} \bibinfo{person}{Hidetoshi Nishimori}.} \bibinfo{year}{1998}\natexlab{}.
\newblock \showarticletitle{Quantum annealing in the transverse Ising model}.
\newblock \bibinfo{journal}{\emph{Physical Review E}}  \bibinfo{volume}{58} (\bibinfo{year}{1998}), \bibinfo{pages}{5355}.
\newblock


\bibitem[Ke et~al\mbox{.}(2025)]%
        {ke2025large}
\bibfield{author}{\bibinfo{person}{Wenjun Ke}, \bibinfo{person}{Yifan Zheng}, \bibinfo{person}{Yining Li}, \bibinfo{person}{Hengyuan Xu}, \bibinfo{person}{Dong Nie}, \bibinfo{person}{Peng Wang}, {and} \bibinfo{person}{Yao He}.} \bibinfo{year}{2025}\natexlab{}.
\newblock \showarticletitle{Large language models in document intelligence: A comprehensive survey, recent advances, challenges, and future trends}.
\newblock \bibinfo{journal}{\emph{ACM Transactions on Information Systems}} \bibinfo{volume}{44}, \bibinfo{number}{1} (\bibinfo{year}{2025}), \bibinfo{pages}{1--64}.
\newblock


\bibitem[Kennedy and Eberhart(1995)]%
        {kennedy1995pso}
\bibfield{author}{\bibinfo{person}{James Kennedy} {and} \bibinfo{person}{Russell Eberhart}.} \bibinfo{year}{1995}\natexlab{}.
\newblock \showarticletitle{Particle swarm optimization}.
\newblock \bibinfo{journal}{\emph{Proceedings of ICNN}} (\bibinfo{year}{1995}), \bibinfo{pages}{1942--1948}.
\newblock


\bibitem[Kim et~al\mbox{.}(2021)]%
        {kim2021donut}
\bibfield{author}{\bibinfo{person}{Geewook Kim}, \bibinfo{person}{Teakgyu Hong}, \bibinfo{person}{Moonbin Yim}, \bibinfo{person}{Jinyoung Park}, \bibinfo{person}{Jinyeong Yim}, \bibinfo{person}{Wonseok Hwang}, \bibinfo{person}{Sangdoo Yun}, \bibinfo{person}{Dongyoon Han}, {and} \bibinfo{person}{Seunghyun Park}.} \bibinfo{year}{2021}\natexlab{}.
\newblock \showarticletitle{Donut: Document understanding transformer without ocr}.
\newblock \bibinfo{journal}{\emph{arXiv preprint arXiv:2111.15664}} \bibinfo{volume}{7}, \bibinfo{number}{15} (\bibinfo{year}{2021}), \bibinfo{pages}{2}.
\newblock


\bibitem[Kirkpatrick et~al\mbox{.}(1983)]%
        {kirkpatrick1983annealing}
\bibfield{author}{\bibinfo{person}{Scott Kirkpatrick}, \bibinfo{person}{C.~Daniel Gelatt}, {and} \bibinfo{person}{Mario~P. Vecchi}.} \bibinfo{year}{1983}\natexlab{}.
\newblock \showarticletitle{Optimization by Simulated Annealing}.
\newblock \bibinfo{journal}{\emph{Science}} \bibinfo{volume}{220}, \bibinfo{number}{4598} (\bibinfo{year}{1983}), \bibinfo{pages}{671--680}.
\newblock


\bibitem[Kitai et~al\mbox{.}(2020)]%
        {kitai2020metamaterials}
\bibfield{author}{\bibinfo{person}{Koki Kitai}, \bibinfo{person}{Jun Guo}, \bibinfo{person}{Shuo Ju}, \bibinfo{person}{Shu Tanaka}, \bibinfo{person}{Koji Tsuda}, \bibinfo{person}{Junichiro Shiomi}, {and} \bibinfo{person}{Ryotaro Tamura}.} \bibinfo{year}{2020}\natexlab{}.
\newblock \showarticletitle{Designing metamaterials with quantum annealing and factorization machines}.
\newblock \bibinfo{journal}{\emph{Physical Review Research}} \bibinfo{volume}{2}, \bibinfo{number}{1} (\bibinfo{year}{2020}), \bibinfo{pages}{013319}.
\newblock


\bibitem[Lee et~al\mbox{.}(2023)]%
        {lee2023pix2struct}
\bibfield{author}{\bibinfo{person}{Kenton Lee}, \bibinfo{person}{Mandar Joshi}, \bibinfo{person}{Iulia~Raluca Turc}, \bibinfo{person}{Hexiang Hu}, \bibinfo{person}{Fangyu Liu}, \bibinfo{person}{Julian~Martin Eisenschlos}, \bibinfo{person}{Urvashi Khandelwal}, \bibinfo{person}{Peter Shaw}, \bibinfo{person}{Ming-Wei Chang}, {and} \bibinfo{person}{Kristina Toutanova}.} \bibinfo{year}{2023}\natexlab{}.
\newblock \showarticletitle{Pix2struct: Screenshot parsing as pretraining for visual language understanding}. In \bibinfo{booktitle}{\emph{International Conference on Machine Learning}}. PMLR, \bibinfo{pages}{18893--18912}.
\newblock


\bibitem[Li et~al\mbox{.}(2021)]%
        {8948239}
\bibfield{author}{\bibinfo{person}{Jianan Li}, \bibinfo{person}{Jimei Yang}, \bibinfo{person}{Aaron Hertzmann}, \bibinfo{person}{Jianming Zhang}, {and} \bibinfo{person}{Tingfa Xu}.} \bibinfo{year}{2021}\natexlab{}.
\newblock \showarticletitle{LayoutGAN: Synthesizing Graphic Layouts With Vector-Wireframe Adversarial Networks}.
\newblock \bibinfo{journal}{\emph{IEEE Transactions on Pattern Analysis and Machine Intelligence}} \bibinfo{volume}{43}, \bibinfo{number}{7} (\bibinfo{year}{2021}), \bibinfo{pages}{2388--2399}.
\newblock
\href{https://doi.org/10.1109/TPAMI.2019.2963663}{doi:\nolinkurl{10.1109/TPAMI.2019.2963663}}


\bibitem[McMinn(2011)]%
        {mcminn2011search}
\bibfield{author}{\bibinfo{person}{Phil McMinn}.} \bibinfo{year}{2011}\natexlab{}.
\newblock \showarticletitle{Search-Based Software Testing: Past, Present and Future}. In \bibinfo{booktitle}{\emph{2011 IEEE Fourth International Conference on Software Testing, Verification and Validation Workshops}}. \bibinfo{pages}{153--163}.
\newblock
\href{https://doi.org/10.1109/ICSTW.2011.100}{doi:\nolinkurl{10.1109/ICSTW.2011.100}}


\bibitem[Mouret and Clune(2015)]%
        {mouret2015mapelites}
\bibfield{author}{\bibinfo{person}{Jean-Baptiste Mouret} {and} \bibinfo{person}{Jeff Clune}.} \bibinfo{year}{2015}\natexlab{}.
\newblock \showarticletitle{Illuminating search spaces by mapping elites}.
\newblock \bibinfo{journal}{\emph{arXiv preprint arXiv:1504.04909}} (\bibinfo{year}{2015}).
\newblock


\bibitem[Nguyen~Tien and Le(2025)]%
        {zhang2024robustness}
\bibfield{author}{\bibinfo{person}{Dong Nguyen~Tien} {and} \bibinfo{person}{Dung~D. Le}.} \bibinfo{year}{2025}\natexlab{}.
\newblock \showarticletitle{Robustness Evaluation of OCR-based Visual Document Understanding under Multi-Modal Adversarial Attacks}.
\newblock \bibinfo{journal}{\emph{arXiv preprint arXiv:2506.16407}} (\bibinfo{year}{2025}).
\newblock


\bibitem[Petke et~al\mbox{.}(2015)]%
        {petke2015practical}
\bibfield{author}{\bibinfo{person}{Justyna Petke}, \bibinfo{person}{Myra~B Cohen}, \bibinfo{person}{Mark Harman}, {and} \bibinfo{person}{Shin Yoo}.} \bibinfo{year}{2015}\natexlab{}.
\newblock \showarticletitle{Practical combinatorial interaction testing: Empirical findings on efficiency and early fault detection}.
\newblock \bibinfo{journal}{\emph{IEEE Transactions on Software Engineering}} \bibinfo{volume}{41}, \bibinfo{number}{9} (\bibinfo{year}{2015}), \bibinfo{pages}{901--924}.
\newblock


\bibitem[Pintore et~al\mbox{.}(2025)]%
        {pintore2024counterfeit}
\bibfield{author}{\bibinfo{person}{Marco Pintore}, \bibinfo{person}{Maura Pintor}, \bibinfo{person}{Dimosthenis Karatzas}, {and} \bibinfo{person}{Battista Biggio}.} \bibinfo{year}{2025}\natexlab{}.
\newblock \showarticletitle{Counterfeit Answers: Adversarial Forgery against OCR-Free Document Visual Question Answering}.
\newblock \bibinfo{journal}{\emph{arXiv preprint arXiv:2512.04554}} (\bibinfo{year}{2025}).
\newblock
\href{https://doi.org/10.48550/arXiv.2512.04554}{doi:\nolinkurl{10.48550/arXiv.2512.04554}}


\bibitem[Pugh et~al\mbox{.}(2016)]%
        {pugh2016qd}
\bibfield{author}{\bibinfo{person}{Justin~K. Pugh}, \bibinfo{person}{Lisa~B. Soros}, {and} \bibinfo{person}{Kenneth~O. Stanley}.} \bibinfo{year}{2016}\natexlab{}.
\newblock \showarticletitle{Quality Diversity: A New Frontier for Evolutionary Computation}.
\newblock \bibinfo{journal}{\emph{Frontiers in Robotics and AI}}  \bibinfo{volume}{3} (\bibinfo{year}{2016}), \bibinfo{pages}{40}.
\newblock
\href{https://doi.org/10.3389/frobt.2016.00040}{doi:\nolinkurl{10.3389/frobt.2016.00040}}


\bibitem[Rendle(2010)]%
        {rendle2010fm}
\bibfield{author}{\bibinfo{person}{Steffen Rendle}.} \bibinfo{year}{2010}\natexlab{}.
\newblock \showarticletitle{Factorization Machines}. In \bibinfo{booktitle}{\emph{2010 IEEE International Conference on Data Mining}}.
\newblock


\bibitem[Research(2023)]%
        {google2023advances}
\bibfield{author}{\bibinfo{person}{Google Research}.} \bibinfo{year}{2023}\natexlab{}.
\newblock \bibinfo{title}{Advances in Document Understanding}.
\newblock
\newblock
\shownote{Online Article on Research Blog highlighting benchmark vs real-world performance}.


\bibitem[Schreiber et~al\mbox{.}(2017)]%
        {schreiber2017deepdesrt}
\bibfield{author}{\bibinfo{person}{Sebastian Schreiber}, \bibinfo{person}{Stefan Agne}, \bibinfo{person}{Ivo Wolf}, \bibinfo{person}{Andreas Dengel}, {and} \bibinfo{person}{Sheraz Ahmed}.} \bibinfo{year}{2017}\natexlab{}.
\newblock \showarticletitle{Deepdesrt: Deep learning for detection and structure recognition of tables in document images}. In \bibinfo{booktitle}{\emph{2017 14th IAPR international conference on document analysis and recognition (ICDAR)}}, Vol.~\bibinfo{volume}{1}. IEEE, \bibinfo{pages}{1162--1167}.
\newblock


\bibitem[Schulman et~al\mbox{.}(2017)]%
        {schulman2017ppo}
\bibfield{author}{\bibinfo{person}{John Schulman}, \bibinfo{person}{Filip Wolski}, \bibinfo{person}{Prafulla Dhariwal}, \bibinfo{person}{Alec Radford}, {and} \bibinfo{person}{Oleg Klimov}.} \bibinfo{year}{2017}\natexlab{}.
\newblock \showarticletitle{Proximal Policy Optimization Algorithms}.
\newblock \bibinfo{journal}{\emph{arXiv preprint arXiv:1707.06347}} (\bibinfo{year}{2017}).
\newblock


\bibitem[Smith(2007)]%
        {smith2007tesseract}
\bibfield{author}{\bibinfo{person}{Ray Smith}.} \bibinfo{year}{2007}\natexlab{}.
\newblock \showarticletitle{An Overview of the Tesseract {OCR} Engine}. In \bibinfo{booktitle}{\emph{Proceedings of the Ninth International Conference on Document Analysis and Recognition (ICDAR)}}.
\newblock


\bibitem[Snoek et~al\mbox{.}(2012)]%
        {snoek2012practical}
\bibfield{author}{\bibinfo{person}{Jasper Snoek}, \bibinfo{person}{Hugo Larochelle}, {and} \bibinfo{person}{Ryan~P Adams}.} \bibinfo{year}{2012}\natexlab{}.
\newblock \showarticletitle{Practical bayesian optimization of machine learning algorithms}.
\newblock \bibinfo{journal}{\emph{Advances in neural information processing systems}}  \bibinfo{volume}{25} (\bibinfo{year}{2012}).
\newblock


\bibitem[Srinivas et~al\mbox{.}(2010)]%
        {srinivas2010gpucb}
\bibfield{author}{\bibinfo{person}{Niranjan Srinivas}, \bibinfo{person}{Andreas Krause}, \bibinfo{person}{Sham Kakade}, {and} \bibinfo{person}{Matthias Seeger}.} \bibinfo{year}{2010}\natexlab{}.
\newblock \showarticletitle{Gaussian Process Optimization in the Bandit Setting}.
\newblock \bibinfo{journal}{\emph{ICML}} (\bibinfo{year}{2010}).
\newblock


\bibitem[Williams(1992)]%
        {williams1992reinforce}
\bibfield{author}{\bibinfo{person}{Ronald~J. Williams}.} \bibinfo{year}{1992}\natexlab{}.
\newblock \showarticletitle{Simple statistical gradient-following algorithms for connectionist reinforcement learning}.
\newblock \bibinfo{journal}{\emph{Machine Learning}}  \bibinfo{volume}{8} (\bibinfo{year}{1992}), \bibinfo{pages}{229--256}.
\newblock


\bibitem[Zanibbi et~al\mbox{.}(2004)]%
        {zanibbi2004survey}
\bibfield{author}{\bibinfo{person}{Richard Zanibbi}, \bibinfo{person}{Dorothea Blostein}, {and} \bibinfo{person}{James~R Cordy}.} \bibinfo{year}{2004}\natexlab{}.
\newblock \showarticletitle{A survey of table recognition: Models, observations, transformations, and inferences}.
\newblock \bibinfo{journal}{\emph{Document Analysis and Recognition}} \bibinfo{volume}{7}, \bibinfo{number}{1} (\bibinfo{year}{2004}), \bibinfo{pages}{1--16}.
\newblock


\bibitem[Zohdinasab et~al\mbox{.}(2023b)]%
        {zohdinasab2021deephyperion}
\bibfield{author}{\bibinfo{person}{Tahereh Zohdinasab}, \bibinfo{person}{Vincenzo Riccio}, \bibinfo{person}{Alessio Gambi}, {and} \bibinfo{person}{Paolo Tonella}.} \bibinfo{year}{2023}\natexlab{b}.
\newblock \showarticletitle{Efficient and Effective Feature Space Exploration for Testing Deep Learning Systems}.
\newblock \bibinfo{journal}{\emph{ACM Trans. Softw. Eng. Methodol.}} \bibinfo{volume}{32}, \bibinfo{number}{2}, Article \bibinfo{articleno}{49} (\bibinfo{date}{March} \bibinfo{year}{2023}), \bibinfo{numpages}{38}~pages.
\newblock
\showISSN{1049-331X}
\href{https://doi.org/10.1145/3544792}{doi:\nolinkurl{10.1145/3544792}}


\bibitem[Zohdinasab et~al\mbox{.}(2023a)]%
        {zohdinasab2023deepatash}
\bibfield{author}{\bibinfo{person}{Tahereh Zohdinasab}, \bibinfo{person}{Vincenzo Riccio}, {and} \bibinfo{person}{Paolo Tonella}.} \bibinfo{year}{2023}\natexlab{a}.
\newblock \showarticletitle{DeepAtash: Focused Test Generation for Deep Learning Systems}. In \bibinfo{booktitle}{\emph{Proceedings of the ACM SIGSOFT International Symposium on Software Testing and Analysis (ISSTA)}}.
\newblock
\href{https://doi.org/10.1145/3597926.3598109}{doi:\nolinkurl{10.1145/3597926.3598109}}


\end{thebibliography}
\appendix
\section{Appendix}
\clearpage
\begin{table*}[h!]
\centering
\caption{Search strategies evaluated under a fixed budget of 1000 oracle calls. All methods use identical batch sizes and initialization where applicable. Each method is executed as an isolated process with a maximum allocation of 64\,GB RAM and 16 CPU cores. Observed runtime variations within same solver (e.g. GA-Exploit vs Explore) arise from our caching mechanism, which eliminates redundant processing costs by skipping the generation pipeline whenever a solver revisits a known configuration $z$. The extended time for quantum is due to the use of CPU simulator, the time taken is expected to greatly reduce in actual hardware. }
\small
\begin{tabular}{l l l c c}
\toprule
\textbf{ID} & \textbf{Method} & \textbf{Key Parameters} & \textbf{Time (24D)} & \textbf{Time (27D)} \\
\midrule
1 & GA (Exploration) &
Pop=50, $p_c=0.9$, $p_m=0.1$, tourn=2 & 5 min & 23 min \\
2 & GA (Exploitation) &
Pop=50, $p_c=0.6$, $p_m=0.01$, tourn=7 & 4 min & 12 min \\
3 & SA &
$T_0=3.0$, $T_f=0.05$, geometric cooling & 25 min & 34 min \\
4 & TPE (Optuna) &
$n_{\text{init}}=100$, top-$k$=50 & 24 min & 30 min \\
5 & GP-EI &
Squared-exp kernel, EI, top-$k$=50 & 52 min & 72 min \\
6 & GP-UCB &
Squared-exp kernel, UCB, top-$k$=50 & 64 min & 76 min \\
7 & PPO (Risk) &
Reward = risk only & 31 min & 47 min \\
8 & PPO (Diversity) &
Risk + entropy bonus ($\beta=0.03$) & 35 min & 59 min \\
9 & REINFORCE &
Policy-gradient with baseline & 61 min & 52 min \\
10 & PSO (Binary) &
$w=0.7$, $c_1=c_2=1.5$ & 23 min & 35 min \\
11 & MAP-Elites &
Archive $(25 \times 25)$, $\sigma=0.05$ & 16 min & 39 min \\
12 & QAOA &
FM surrogate, standard mixer & 69 min & 613 min \\
13 & QAOA-Corr &
FM surrogate, correlated mixer & 77 min & 719 min \\
14 & Random &
Uniform sampling & 36 min & 83 min \\
\bottomrule
\end{tabular}
\label{tab:methods}
\end{table*}

\begin{figure*}[h]
    \centering
    \includegraphics[width=0.6\linewidth]{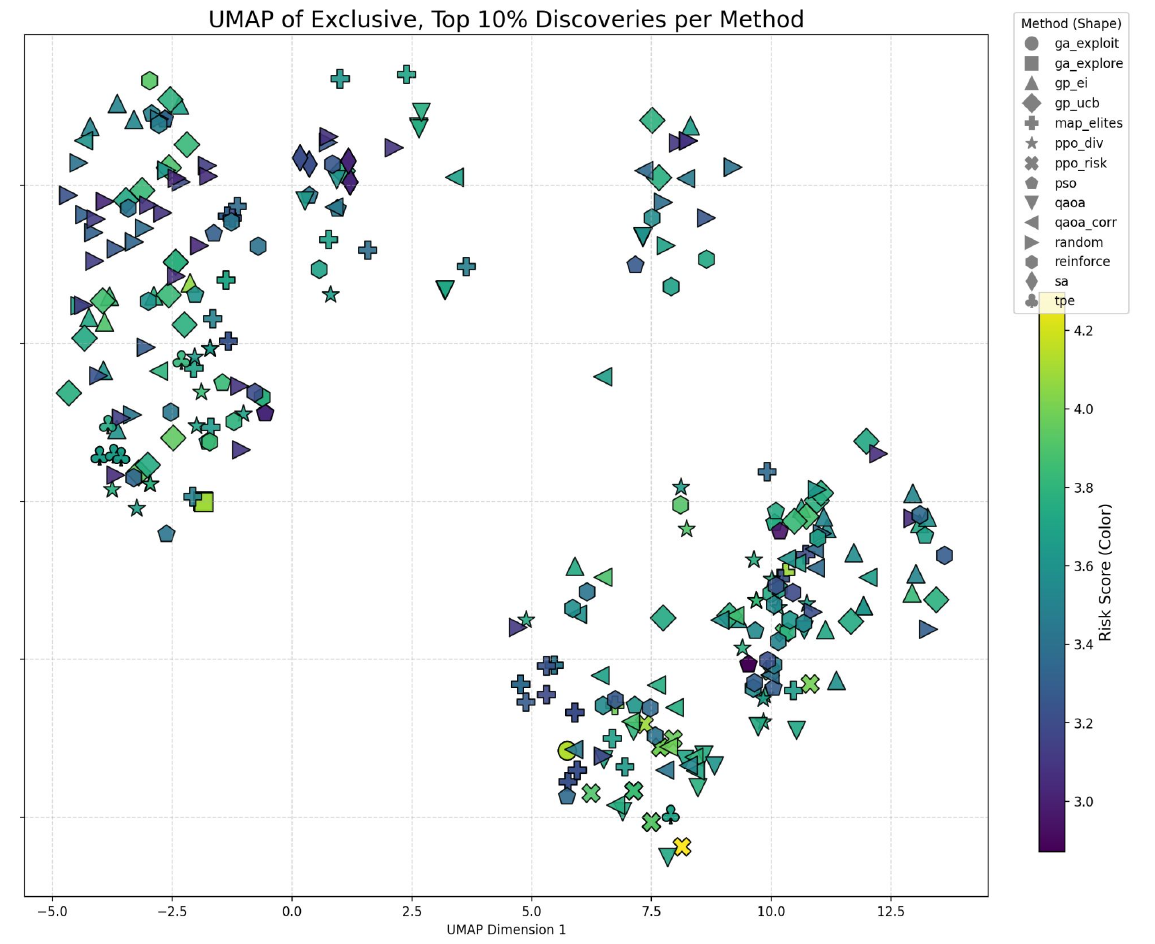}
    \vspace{-0.3cm}
    \caption{
    UMap visualization of exclusive discoveries per method. Points represent "Crown Jewel" samples—unique discoveries within the top 10\% of risk. The plot highlights the unique contributions of each solver, with diversity-oriented methods (e.g., GA-Explore: 44, MAP-Elites: 42) and interaction-aware explorers (QAOA-Corr: 28) identifying distinct failure modes missed by local search baselines (SA: 19, TPE: 21).
    }
    \label{fig:umap_top_exclusive}
\end{figure*}

\begin{figure*}[h]
    \centering
    \includegraphics[width=0.9\linewidth]{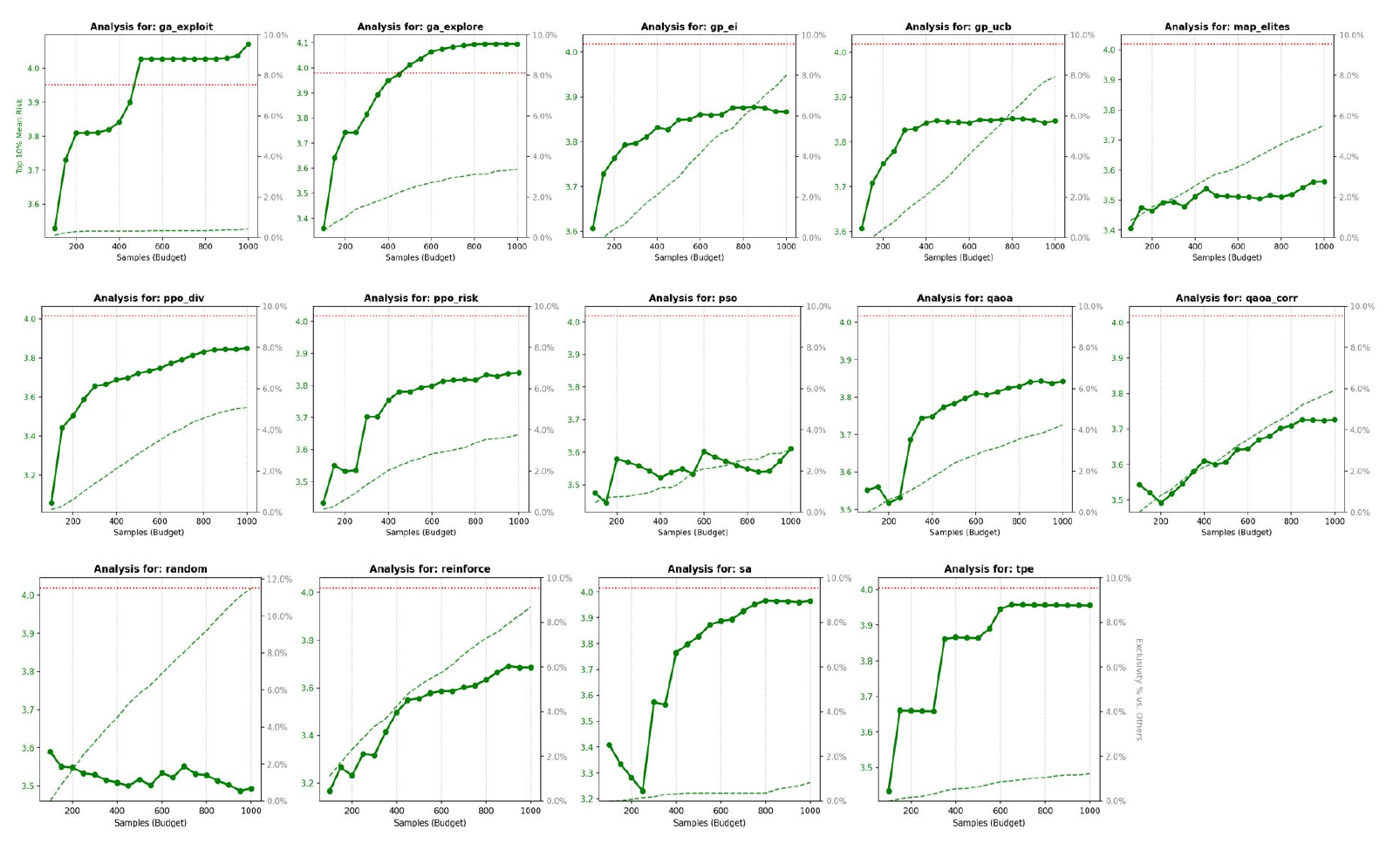}
    \vspace{-0.3cm}
    \caption{
    Configuration-level exclusivity over budget for all methods in single-page (24D).
    Each curve reports the fraction of configurations exclusively discovered by a method relative to all others across same budget. The dotted red line represents the top 10\% mean risk over all other methods, while the dashed green line represents its top 10\% mean risk per budget.
    }
    \label{fig:exclusivity_over_budget}
\end{figure*}


\clearpage
\paragraph{\textbf{Pairwise Exclusive Risk Modes: QAOA-Corr vs. REINFORCE}}
\label{sec:qaoa_vs_reinforce}
We provide an exhaustive enumeration of the risk modes discovered by \textbf{QAOA-Corr} (Total: 26) and \textbf{REINFORCE} (Total: 24). The intersection of these two strategies yields \textbf{23 shared modes}, confirming a strong consensus on the dominant failure attractors. However, distinct exploration behaviors emerge in the exclusive sets: QAOA-Corr isolates 3 unique modes in the medium-density regime, while REINFORCE identifies 1 unique mode in the low-density regime.

\paragraph{Intersection: Common Risk Modes (23)}
The following modes were successfully identified by \textbf{both} strategies:
\begin{itemize} \setlength\itemsep{0em} \footnotesize
    \item \texttt{DENSITY:HIGH | FAILURE:SUMMARY\_MISSING | FAILURE:TABLE\_TRUNCATED | LAYOUT:HARD\_SPLIT}
    \item \texttt{DENSITY:LOW | FAILURE:SUMMARY\_MISSING | FAILURE:TABLE\_TRUNCATED | LAYOUT:HARD\_SPLIT}
    \item \texttt{DENSITY:LOW | FAILURE:SUMMARY\_MISSING | LAYOUT:HARD\_SPLIT}
    \item \texttt{DENSITY:LOW | FAILURE:SUMMARY\_MISSING | FAILURE:TABLE\_TRUNCATED | LAYOUT:NO\_SPLIT}
    \item \texttt{DENSITY:LOW | FAILURE:TABLE\_TRUNCATED | LAYOUT:HARD\_SPLIT}
    \item \texttt{DENSITY:LOW | LAYOUT:HARD\_SPLIT}
    \item \texttt{DENSITY:LOW | FAILURE:TABLE\_TRUNCATED | LAYOUT:NO\_SPLIT}
    \item \texttt{DENSITY:LOW | LAYOUT:NO\_SPLIT}
    \item \texttt{DENSITY:LOW | FAILURE:SUMMARY\_MISSING | LAYOUT:SOFT\_SPLIT}
    \item \texttt{DENSITY:LOW | FAILURE:SUMMARY\_MISSING | FAILURE:TABLE\_TRUNCATED | LAYOUT:SOFT\_SPLIT}
    \item \texttt{DENSITY:LOW | FAILURE:TABLE\_TRUNCATED | LAYOUT:SOFT\_SPLIT}
    \item \texttt{DENSITY:LOW | LAYOUT:SOFT\_SPLIT}
    \item \texttt{DENSITY:MEDIUM | FAILURE:SUMMARY\_MISSING | FAILURE:TABLE\_TRUNCATED | LAYOUT:NO\_SPLIT}
    \item \texttt{DENSITY:MEDIUM | FAILURE:SUMMARY\_MISSING | FAILURE:TABLE\_TRUNCATED | LAYOUT:HARD\_SPLIT}
    \item \texttt{DENSITY:MEDIUM | FAILURE:TABLE\_TRUNCATED | LAYOUT:HARD\_SPLIT}
    \item \texttt{DENSITY:MEDIUM | FAILURE:TABLE\_TRUNCATED | LAYOUT:NO\_SPLIT}
    \item \texttt{DENSITY:MEDIUM | FAILURE:SUMMARY\_MISSING | FAILURE:TABLE\_TRUNCATED | LAYOUT:SOFT\_SPLIT}
    \item \texttt{DENSITY:MEDIUM | FAILURE:TABLE\_TRUNCATED | LAYOUT:SOFT\_SPLIT}
    \item \texttt{DENSITY:HIGH | FAILURE:SUMMARY\_MISSING | FAILURE:TABLE\_TRUNCATED | LAYOUT:NO\_SPLIT}
    \item \texttt{DENSITY:HIGH | FAILURE:TABLE\_TRUNCATED | LAYOUT:HARD\_SPLIT}
    \item \texttt{DENSITY:HIGH | FAILURE:TABLE\_TRUNCATED | LAYOUT:NO\_SPLIT}
    \item \texttt{DENSITY:HIGH | FAILURE:TABLE\_TRUNCATED | LAYOUT:SOFT\_SPLIT}
    \item \texttt{DENSITY:HIGH | FAILURE:SUMMARY\_MISSING | FAILURE:TABLE\_TRUNCATED | LAYOUT:SOFT\_SPLIT}
\end{itemize}

\paragraph{Exclusive Discoveries}
The divergence in search trajectories is highlighted by the following unique modes:

\textbf{Exclusive to QAOA-Corr (3):}
\begin{itemize} \setlength\itemsep{0em} \footnotesize
    \item \texttt{DENSITY:MEDIUM | LAYOUT:HARD\_SPLIT}
    \item \texttt{DENSITY:MEDIUM | LAYOUT:NO\_SPLIT}
    \item \texttt{DENSITY:MEDIUM | LAYOUT:SOFT\_SPLIT}
\end{itemize}

\textbf{Exclusive to REINFORCE (1):}
\begin{itemize} \setlength\itemsep{0em} \footnotesize
    \item \texttt{DENSITY:LOW | FAILURE:SUMMARY\_MISSING | LAYOUT:NO\_SPLIT}
\end{itemize}

\begin{figure*}[h]
    \centering
    \includegraphics[width=0.9\linewidth]{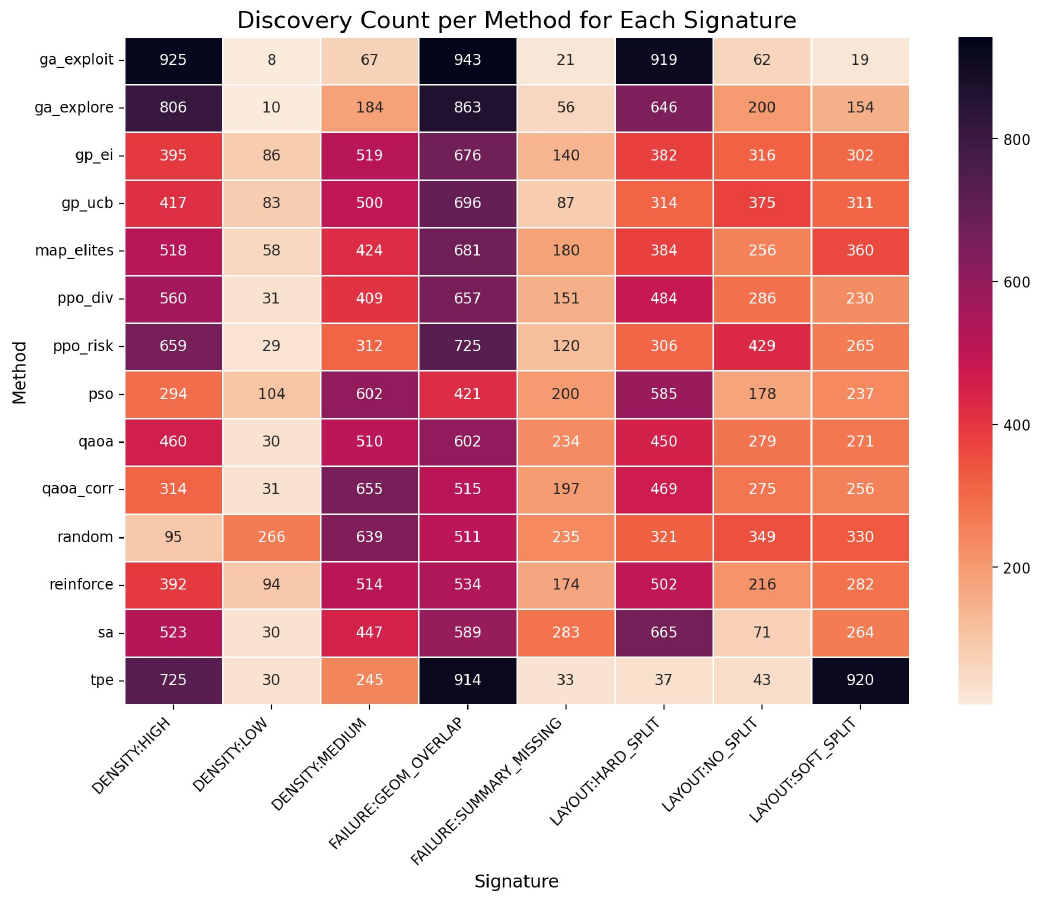}
    \vspace{-0.3cm}
    \caption{
    Count of each risk signature discovered by each method, across same budget of 1000. The signatures are independent and can occur for every sample, hence the total count of all signatures do not necessarily need to add to 1000. On the other hand, risk modes denote the interaction of two or more signatures and are limited in number. 
    }
    \label{fig:risk_signatures}
\end{figure*}

\begin{figure*}[h]
    \centering
    \includegraphics[width=0.9\linewidth]{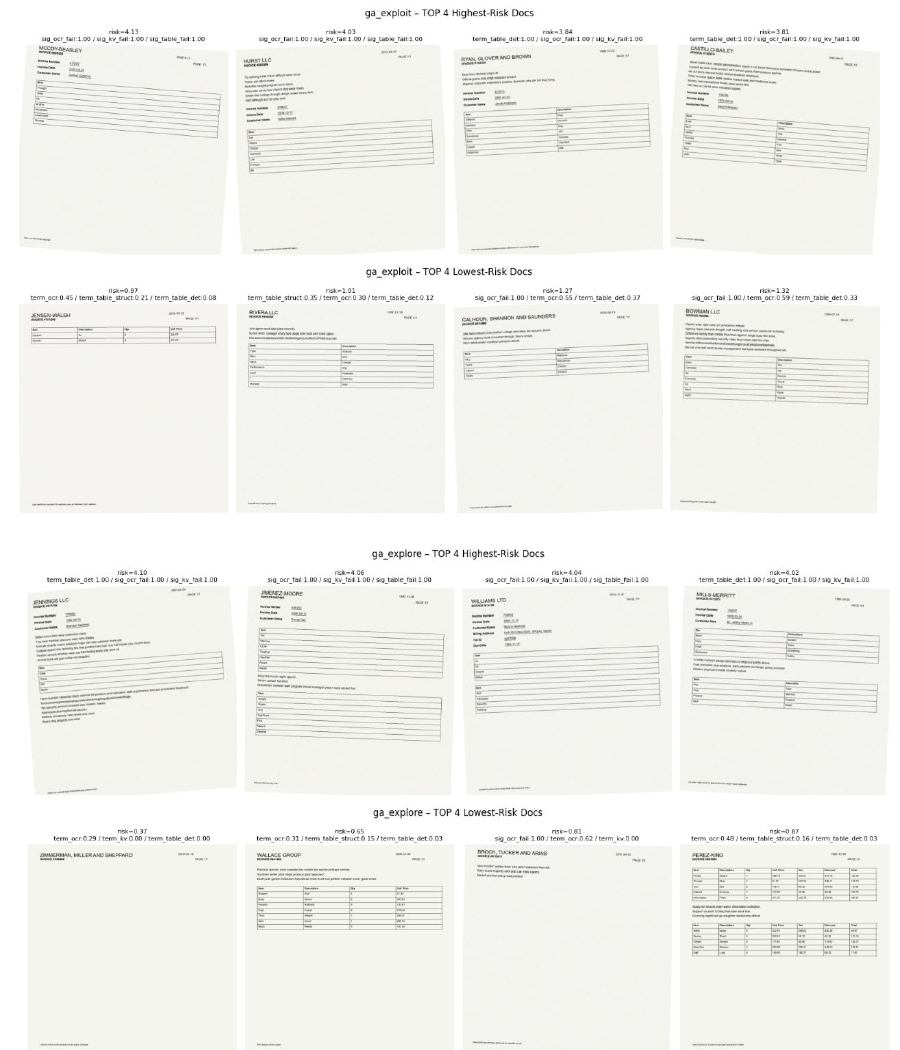}
    \vspace{-0.3cm}
    \caption{
   Visualization of representative high and low-risk synthetic documents discovered by GA, along with its computed risk score and the specific failure signature as evaluated by the IDP oracle.
    }
    \label{fig:doc1}
\end{figure*}

\begin{figure*}[h]
    \centering
    \includegraphics[width=0.9\linewidth]{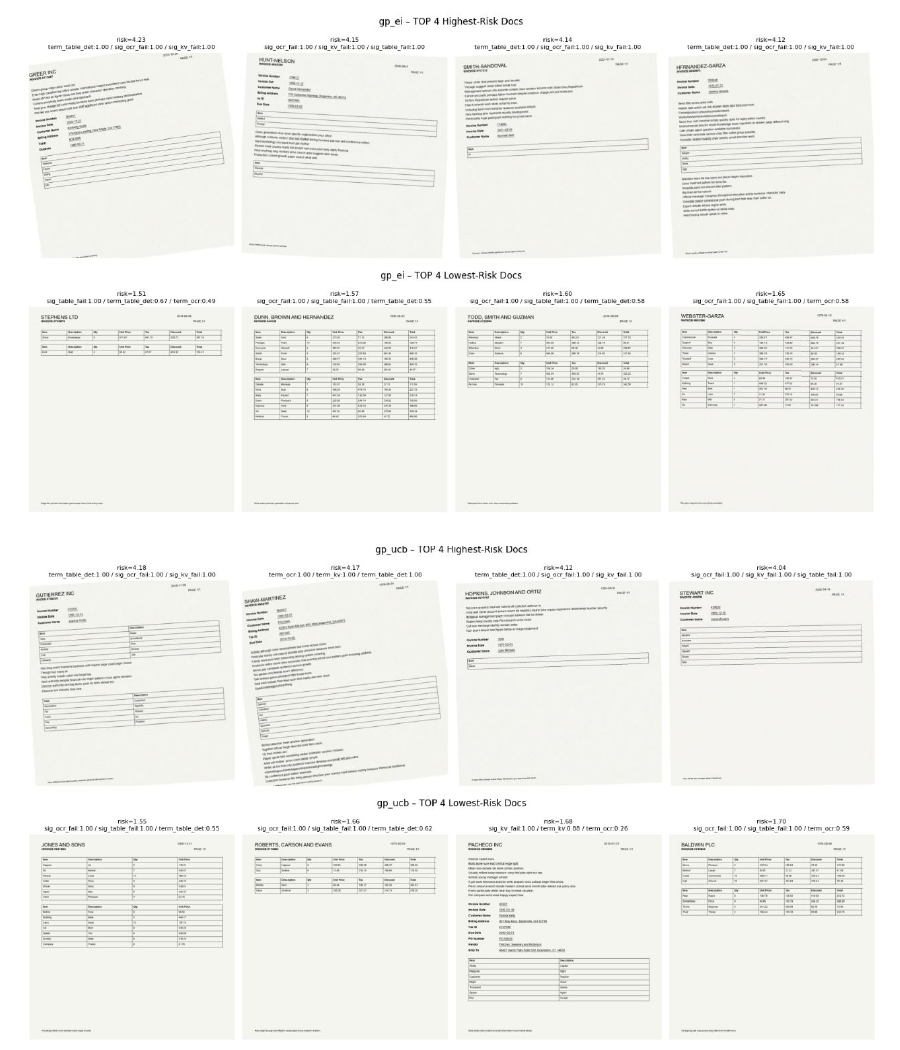}
    \vspace{-0.3cm}
    \caption{
   Visualization of representative high and low-risk synthetic documents discovered by GP (BO), along with its computed risk score and the specific failure signature as evaluated by the IDP oracle.
    }
    \label{fig:doc2}
\end{figure*}

\begin{figure*}[h]
    \centering
    \includegraphics[width=0.9\linewidth]{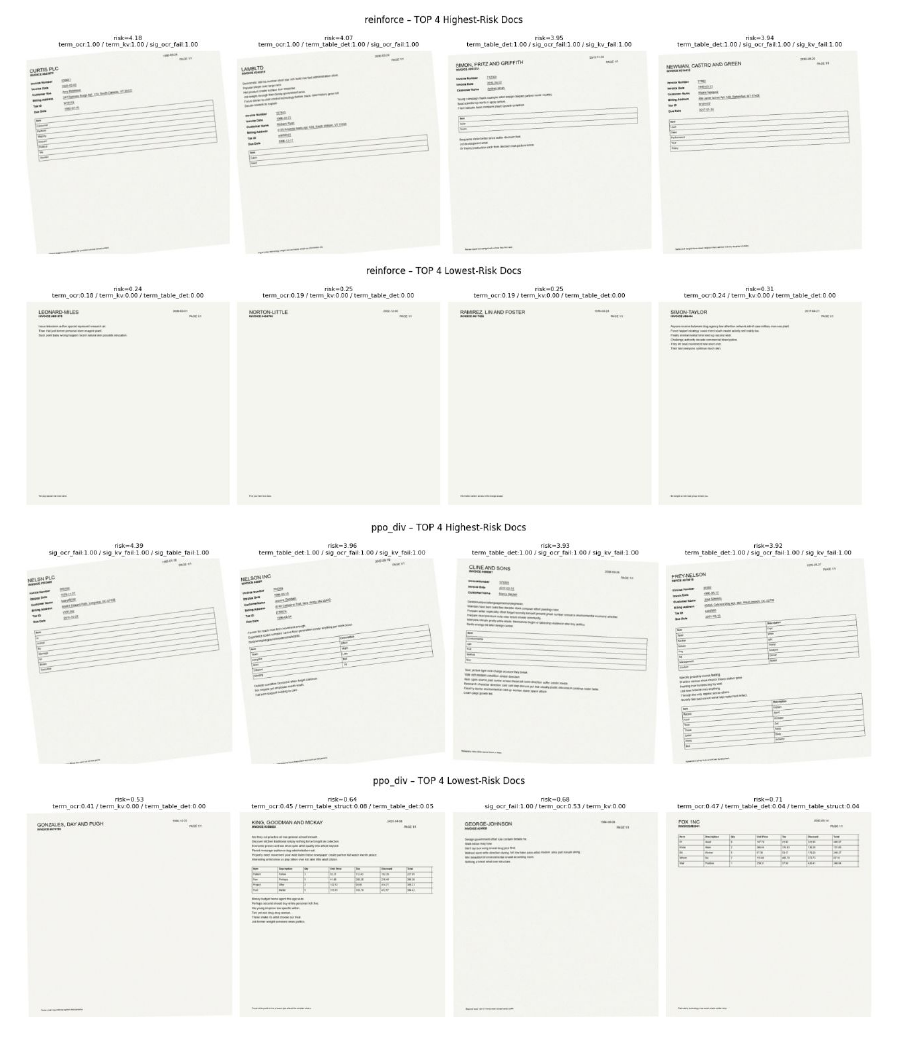}
    \vspace{-0.3cm}
    \caption{
   Visualization of representative high and low-risk synthetic documents discovered by RL, along with its computed risk score and the specific failure signature as evaluated by the IDP oracle.
    }
    \label{fig:doc3}
\end{figure*}

\begin{figure*}[h]
    \centering
    \includegraphics[width=0.9\linewidth]{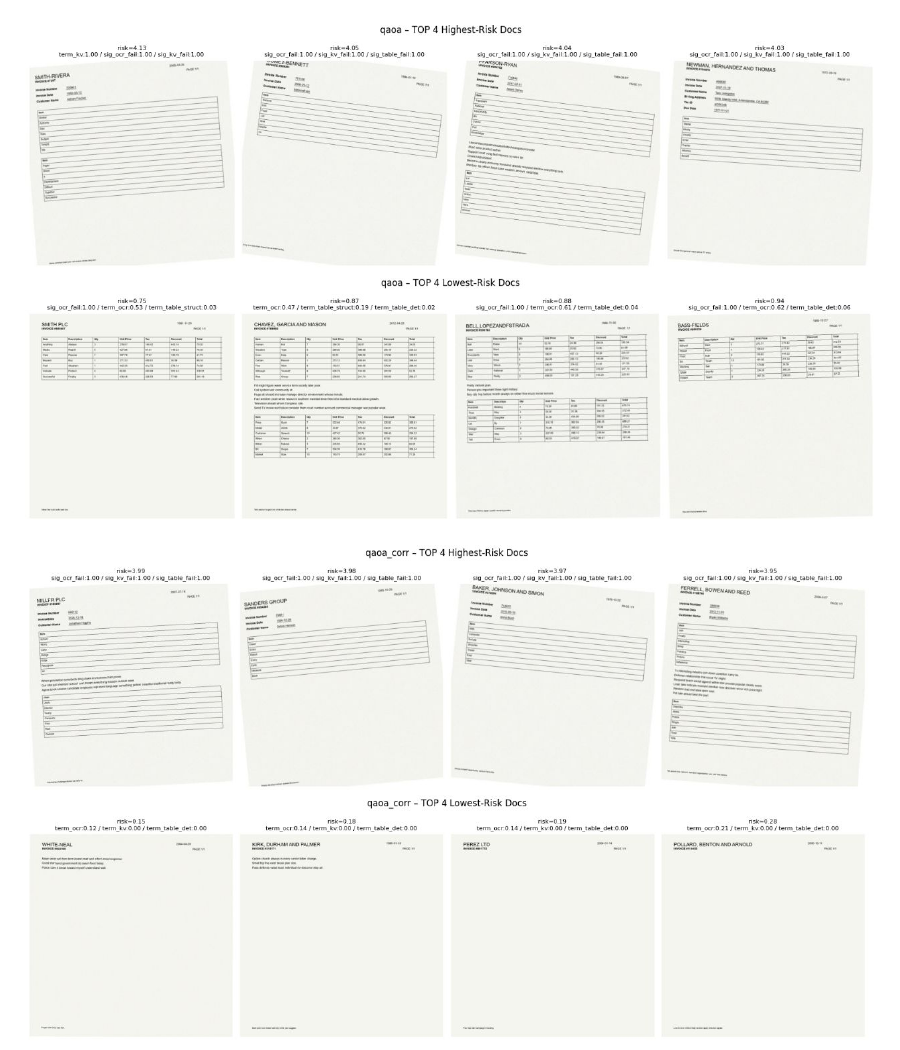}
    \vspace{-0.3cm}
    \caption{
   Visualization of representative high and low-risk synthetic documents discovered by QAOA, along with its computed risk score and the specific failure signature as evaluated by the IDP oracle.
    }
    \label{fig:doc4}
\end{figure*}

\begin{figure*}[h]
    \centering
    \includegraphics[width=0.9\linewidth]{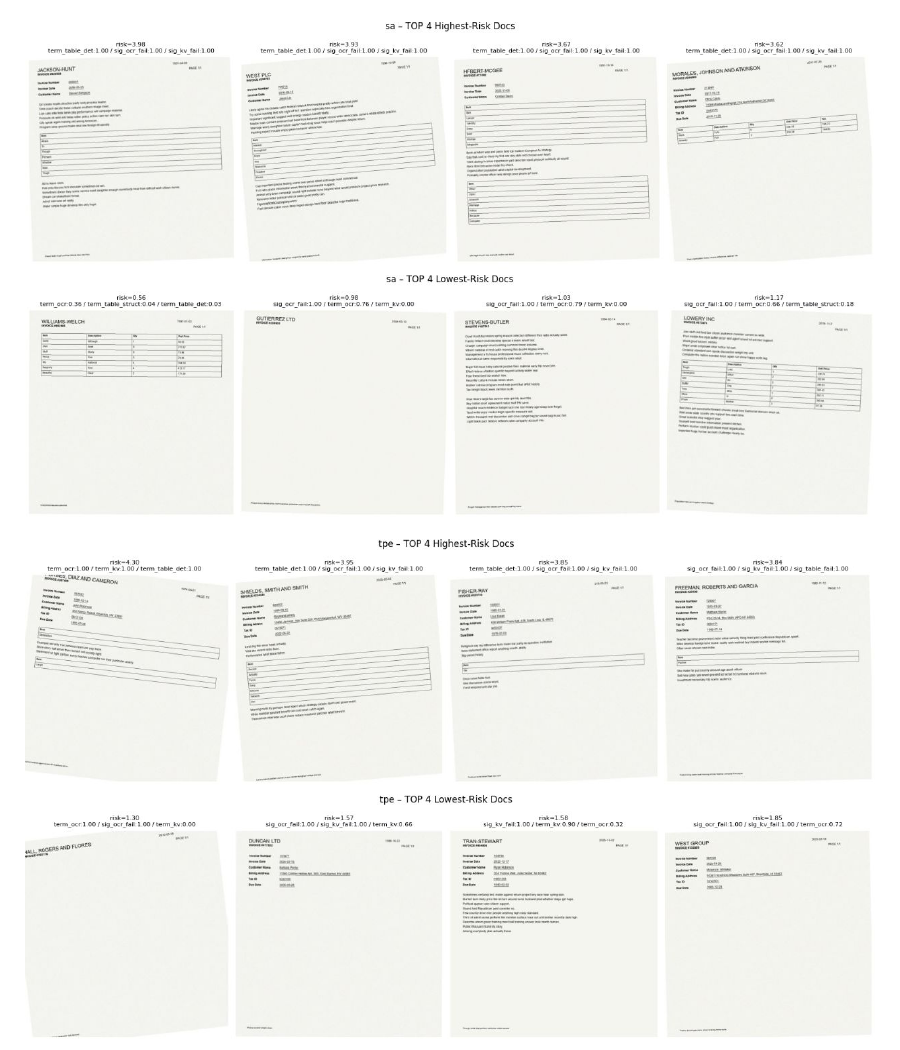}
    \vspace{-0.3cm}
    \caption{
   Visualization of representative high and low-risk synthetic documents discovered by SA and TPE, along with its computed risk score and the specific failure signature as evaluated by the IDP oracle.
    }
    \label{fig:doc5}
\end{figure*}

\begin{figure*}[h]
    \centering
    \includegraphics[width=0.9\linewidth]{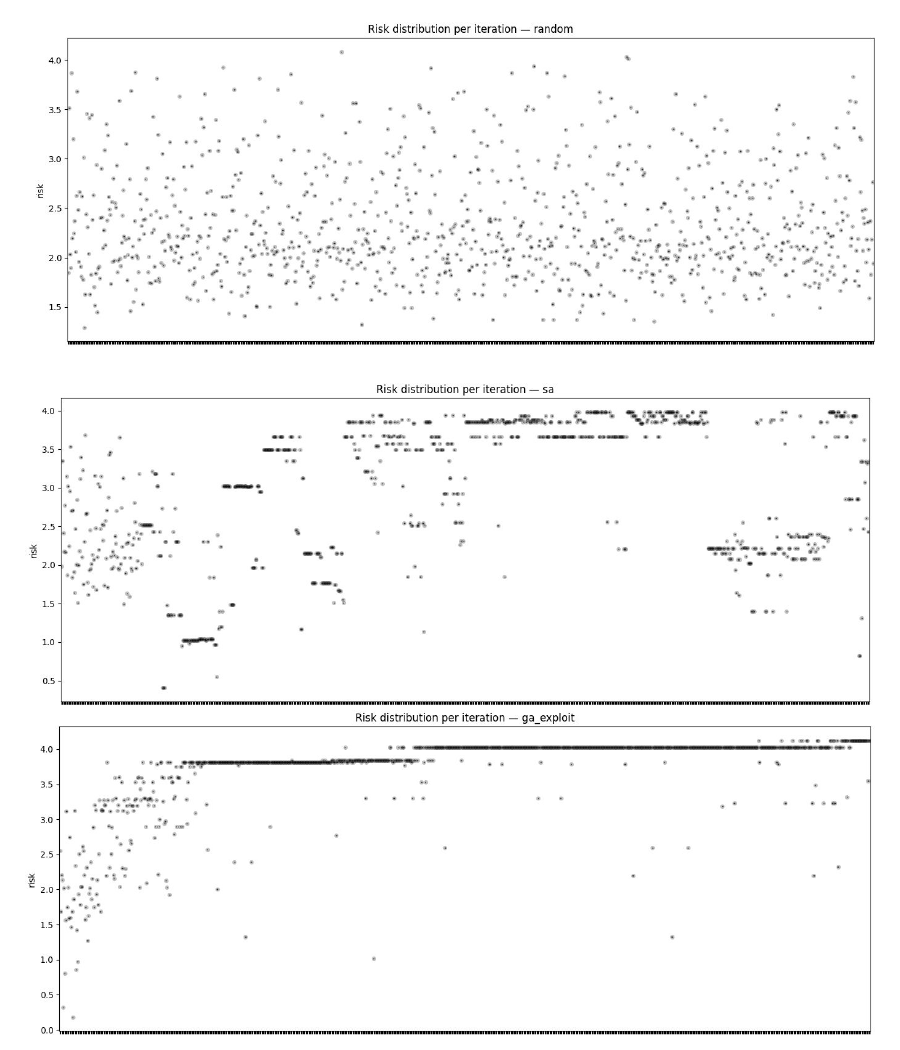}
    \vspace{-0.3cm}
    \caption{
  Scatter plot showing the distribution of risk values ($y$-axis) discovered by each solver as the evaluation budget progresses ($x$-axis).
    }
    \label{fig:risk_z_plot_1}
\end{figure*}

\begin{figure*}[h]
    \centering
    \includegraphics[width=0.9\linewidth]{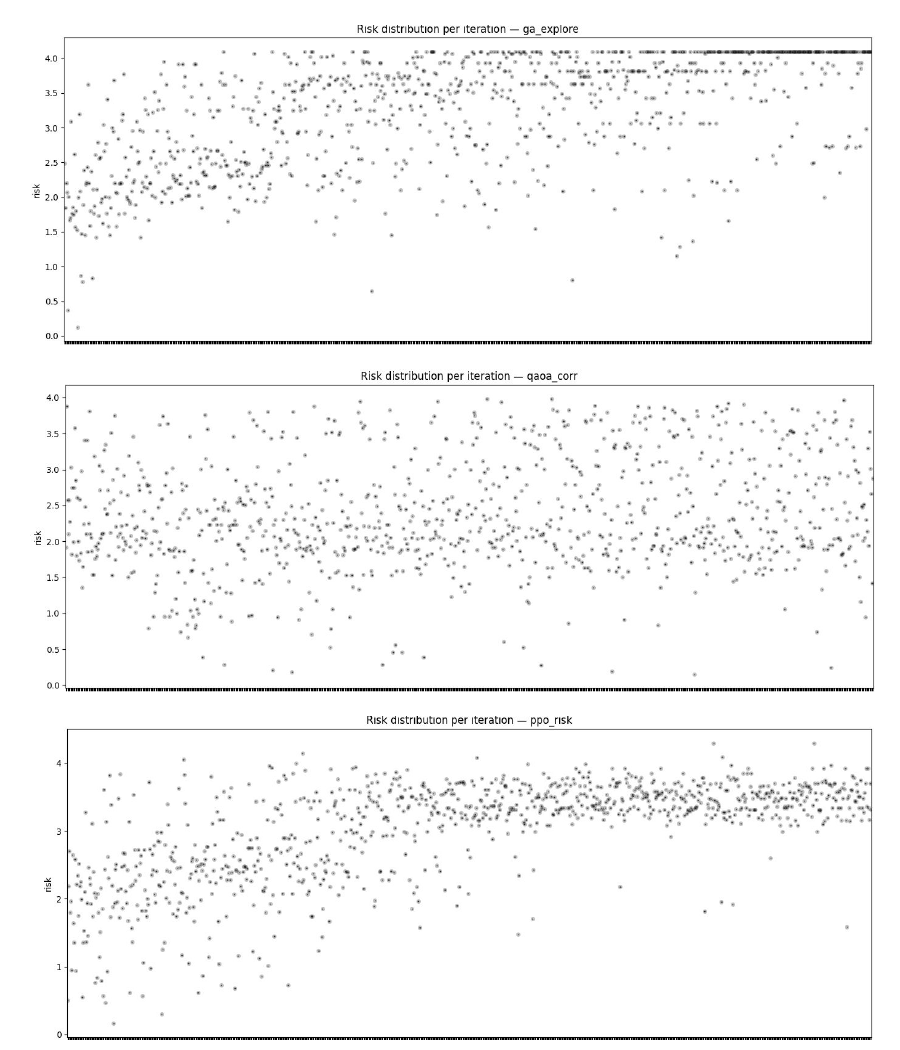}
    \vspace{-0.3cm}
    \caption{
  Scatter plot showing the distribution of risk values ($y$-axis) discovered by each solver as the evaluation budget progresses ($x$-axis).
    }
    \label{fig:risk_z_plot_1}
\end{figure*}

\begin{figure*}[h]
    \centering
    \includegraphics[width=0.9\linewidth]{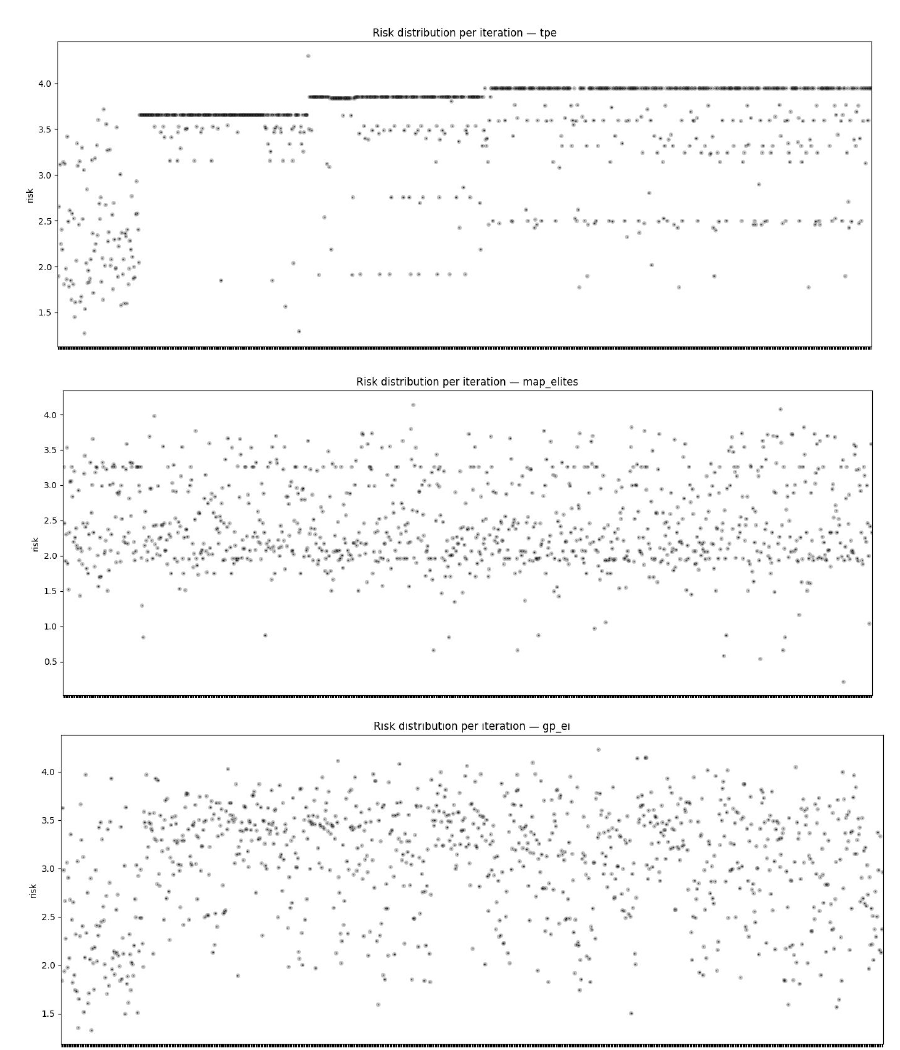}
    \vspace{-0.3cm}
    \caption{
  Scatter plot showing the distribution of risk values ($y$-axis) discovered by each solver as the evaluation budget progresses ($x$-axis).
    }
    \label{fig:risk_z_plot_3}
\end{figure*}

\end{document}